\documentclass[10pt,twocolumn,letterpaper]{article}

\usepackage[pagenumbers]{cvpr} %

\usepackage{pgfplots}

\usepackage{xcolor} %
\definecolor{mypurple}{HTML}{A800FF}
\definecolor{mygray}{HTML}{E6E6E6}

\newcommand{\TODO}[1]{\textbf{\color{red}[TODO: #1]}}
\renewcommand{\TODO}[1]{}

\makeatletter
\renewcommand\paragraph{\@startsection{paragraph}{4}{\z@}%
    {1.0ex \@plus1ex \@minus.2ex}%
    {-\the\fontdimen2\font}%
    {\normalfont\normalsize\bfseries}}%
\makeatother

\usepackage{stmaryrd}
\usepackage{amsmath}
\usepackage{mathtools}
\def\RR{\mathbb{R}}

\DeclareMathOperator*{\argmin}{arg\,min}

\usepackage{tabularx}
\usepackage{multirow}
\newcolumntype{C}[1]{>{\centering\arraybackslash}p{#1}}	
\newcolumntype{M}[1]{>{\centering\arraybackslash}m{#1}}
\newcolumntype{L}[1]{>{\arraybackslash}p{#1}}

\usepackage{cuted}

\usepackage{colortbl}

\usepackage{xfakebold}
\usepackage{adjustbox}

\usepackage{tikz}
\usetikzlibrary{arrows.meta}
\usetikzlibrary{fit}

\newcommand{\stimes}{\hspace{-0.15em}\times\hspace{-0.15em}}

\usepackage{subcaption}
\usepackage{MnSymbol}

\definecolor{cvprblue}{rgb}{0.21,0.49,0.74}
\usepackage[pagebackref,breaklinks,colorlinks,citecolor=cvprblue]{hyperref}

\def\paperID{7984} %
\def\confName{CVPR}
\def\confYear{2024}

\newcommand{\cs}[1]{}

\newcommand{\camera}[1]{#1}
\title{Dense Optical Tracking: Connecting the Dots}
\newcommand{\method}{DOT}
\newcommand{\Method}{Dense optical tracking}

\author{Guillaume Le Moing\textsuperscript{1, 2, }\thanks{corresponding author: guillaume.le-moing@inria.fr} \hspace{0.8cm} Jean Ponce\textsuperscript{2, 3} \hspace{0.8cm} Cordelia Schmid\textsuperscript{1, 2}\\[0.5em]
\begin{tabular}{@{}cccc@{}}
\textsuperscript{1}Inria & ~~~~~~\textsuperscript{2}D\'epartement d’informatique de~~~~~~ & \textsuperscript{3}Courant Institute and Center for Data Science\\
& l’ENS (CNRS, ENS-PSL, Inria) &  New York University\\[-0.3em]
\end{tabular}}

\begin{document}
\maketitle
\begin{abstract}
Recent approaches to point tracking are able to recover the trajectory of \textbf{\textit{any}} scene point through a large portion of a video despite the presence of occlusions.
They are, however, too slow  in practice to track \textbf{\textit{every}} point observed in a single frame in a reasonable amount of time.
This paper introduces \method{}, a novel, simple and efficient method for solving this problem.
It first extracts a small set of tracks from key regions at motion boundaries using an off-the-shelf point tracking algorithm.
Given source and target frames, \method{} then computes rough initial estimates of a dense flow field and visibility mask through nearest-neighbor interpolation, before refining them using a learnable optical flow estimator that explicitly handles occlusions and can be trained on synthetic data with ground-truth correspondences.
We show that \method{} is significantly more accurate than current optical flow techniques, outperforms sophisticated ``universal'' trackers like OmniMotion, and is on par with, or better than, the best point tracking algorithms like CoTracker while being at least two orders of magnitude faster.
Quantitative and qualitative experiments with synthetic and real videos validate the promise of the proposed approach.
Code, data, and videos showcasing the capabilities of our approach are available in the project webpage.\footnote{\url{https://16lemoing.github.io/dot}}
\end{abstract}

\section{Introduction}
\label{sec:introduction}

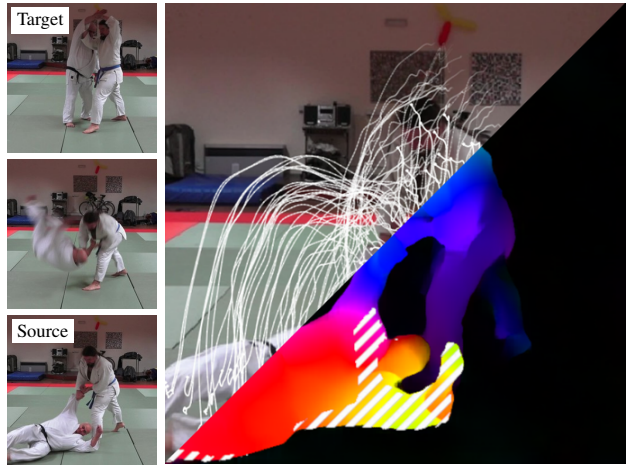
\begin{figure}
\centering

\newcommand{\withText}[2]{%
\begin{tikzpicture}
    \node[anchor=south west, inner sep=0, outer sep=0] (image) at (0,0) {#1};
    
    \node[anchor=north west, text=black] at (-0.15, 2.05) {\setlength{\fboxsep}{2pt} \footnotesize \colorbox{white}{#2}};
\end{tikzpicture}
}
\newcommand{\withTwoTexts}[3]{%
\begin{tikzpicture}
    \node[anchor=south west, inner sep=0, outer sep=0] (image) at (0,0) {#1};
    
    \node[anchor=north west, text=black] at (-0.15, 6.3) {};
    \node[anchor=north west, text=black] at (-0.15, 6.195) {\setlength{\fboxsep}{2pt} \footnotesize \colorbox{white}{#2}};
    \node[anchor=south east, text=black] at (6.175, -0.05) {\setlength{\fboxsep}{2pt} \footnotesize \colorbox{white}{#3}};
\end{tikzpicture}
}

\newcommand{\withoutText}[1]{%
\begin{tikzpicture}
    \node[anchor=south west, inner sep=0, outer sep=0] (image) at (0,0) {#1};

    \node[anchor=north west, text=black] at (-0.15, 2.05) {};
\end{tikzpicture}
}
\hspace{-1em}
\renewcommand{\arraystretch}{0.25}
\begin{tabular}{@{}C{1.98cm}@{}}

\withText{\includegraphics[width=\linewidth]{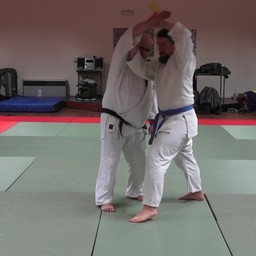}}{\scriptsize Target} \\
\withoutText{\includegraphics[width=\linewidth]{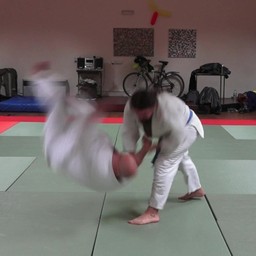}} \\
\withText{\includegraphics[width=\linewidth]{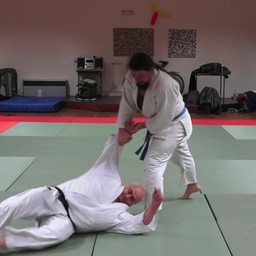}}{\scriptsize Source}
\end{tabular}
\hspace{0.005em}
\renewcommand{\arraystretch}{0.75}
\begin{tabular}{@{}C{6.125cm}@{}}
\\[-0.55em]
\includegraphics[width=\linewidth]{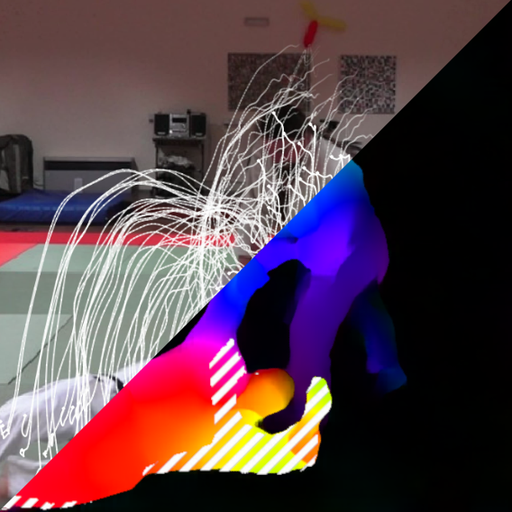}
\end{tabular}
\vspace{-0.5em}

\caption{\textbf{\method{}} unifies point tracking and optical flow techniques. From a few initial tracks, it predicts dense motions and occlusions from source to target frames. We represent tracks in white, occlusions with stripes, and motion directions using distinctive colors.} %
\label{fig:teaser}
\end{figure}

A fine-grain analysis of motion is crucial in many applications involving video data, including frame interpolation~\cite{huang2022rife, li2023amt}, inpainting~\cite{li2022towards, zhou2023propainter}, motion segmentation~\cite{yang2021self, cheng2022implicit}, compression~\cite{agustsson2020scale, lu2019dvc}, future prediction~\cite{wu2020future, le2023waldo}, or editing~\cite{kasten2021layered, ye2022deformable}. 
Historically, most systems designed for these tasks have heavily relied on optical flow algorithms~\cite{teed2020raft, sun2018pwc, dosovitskiy2015flownet} %
with an inherent lack of robustness to %
large motions or occlusions~\cite{fortun2015optical}, confining their use of context to a handful of neighboring frames and limiting long-term reasoning.

Lately, point tracking methods~\cite{doersch2022tap, doersch2023tapir, zheng2023pointodyssey, harley2022particle, karaev2023cotracker} have emerged as a promising alternative. 
They are, in most cases, able to track {\textbf{any}} specific point through a large portion of a video, even in the presence of occlusions, successfully retaining more than 50\% of initial queries after thousands of frames~\cite{zheng2023pointodyssey}. 
Yet, these methods remain too slow and memory intensive to track {\textbf{every}} individual point in a video, that is, to obtain a set of tracks dense enough to cover every pixel location at every time step.
This computational hurdle has thus far limited the widespread adoption of point tracking as a viable replacement for optical flow in downstream tasks. 

\cs{whats the time limitation of these methods? how long is the video, probably not one hour?}
\cs{give examples and expand a bit}

The inefficiency of point tracking methods~\cite{doersch2022tap, doersch2023tapir, zheng2023pointodyssey, harley2022particle} arises from their independent processing of individual tracks. In a recent study~\cite{karaev2023cotracker}, Karaev \etal shed light on a related issue: tracking individual queries independently lacks spatial context. %
Their approach, CoTracker, tracks multiple points together to exploit the correlations among their trajectories.
In some cases, performance drops when the number of simultaneous tracks increases too much, thus preventing solutions which are both fast and accurate.

\cs{is the gain only in time or also in quality, can we discuss this a bit, from our discussion and the results, I also understand that the quality is better}

In this paper, we take this concept one step further by tracking every point in a frame simultaneously. Our approach, \method{}, connects the dots (hence its name) between optical flow and point tracking methods (Figure~\ref{fig:teaser}), enjoying the spatial coherence of the former and the temporal consistency of the latter.
Our contributions are as follows: 
\begin{itemize}
    \item We introduce \method{}, a novel, simple and efficient approach that unifies point tracking and optical flow, using a small set of tracks to predict a dense flow field and a visibility mask between arbitrary frames in a video.
    \item We extend the CVO benchmark~\cite{wu2023accflow} with 500 new videos to enhance the assessment of dense and long-term tracking. The new videos are longer and have a higher frame rate than existing ones, for more challenging motions.
    \item We use extensive experiments on the CVO and TAP benchmarks~\cite{wu2023accflow, doersch2022tap}, with quantitative and qualitative results, to demonstrate that \method{} significantly outperforms state-of-the-art optical flow methods and is on par with, or better than, the best point tracking algorithms while being much faster at dense prediction ($\times 100$ speedup).
\end{itemize}

\section{Related work}
\label{sec:related}

\paragraph{Optical flow estimation} has traditionally been addressed using variational methods~\cite{horn1981determining} to minimize an energy function enforcing spatial smoothness and brightness constancy constraints. %
Several works have focused on adapting this to long-range motions, including coarse-to-fine warping strategies~\cite{anandan1989computational, memin2002hierarchical,papenberg2006highly}, a quadratic relaxation of the original problem~\cite{steinbrucker2009large}, non-local regularization~\cite{krahenbuhl2012efficient, ranftl2014non}, nearest-neighbor fields~\cite{chen2013large, barnes2009patchmatch, bailer2015flow, hu2016efficient},
 and methods relying on local feature matching~\cite{brox2010large, weinzaepfel2013deepflow, revaud2015epicflow, wulff2015efficient, hu2017robust}. %

\method{} builds on these classical methods, also refining dense motions from sparse correspondences, {\em but using point tracking instead of local feature matching}. %
Recently, supervised approaches trained on synthetic data~\cite{butler2012naturalistic, dosovitskiy2015flownet, greff2022kubric, mayer2016large} have emerged as a powerful alternative to variational methods. {\em \method{} also benefits from these advances}.

\paragraph{Optical flow in the deep learning era.} 
FlowNet~\cite{dosovitskiy2015flownet, ilg2017flownet} was the first convolutional neural network used to directly predict optical flow from image pairs.
Several classical ideas have been used to enhance accuracy.
For instance, SPyNet~\cite{ranjan2017optical} incorporates a feature pyramid~\cite{burt1983laplacian}, and DCFlow~\cite{xu2017accurate} uses 4D cost volumes to compute correlations between every pair of patches across successive frames, similar to~\cite{brox2010large}. PWC-Net~\cite{sun2018pwc} combines the two ideas.
\camera{SF-Net~\cite{zhou2019moving} builds upon sparse-to-dense methods.}
RAFT~\cite{teed2020raft} uses a recurrent neural network to progressively refine the flow by iterative look-up in the cost volumes.
Lately, transformer-based approaches~\cite{xu2022gmflow, jiang2021learning, sui2022craft, jaegle2021perceiver, huang2022flowformer} have been used to model long-range dependencies and identify similarities between remote patches in pairs of frames. 

Contrary to these methods that only consider pairs of isolated frames, {\em \method{} takes full videos as input}. 
We show that temporal information helps handling occlusions, enhances robustness in the presence of large displacements, and disentangles the motions of objects with similar visual characteristics, in particular for distant frames.

\paragraph{Optical flow across distant frames} has been approached as an interpolation problem given a sparse set of trajectories extracted from a video~\cite{gibson2003robust}.
Another early attempt~\cite{lim2005optical} involved computing the flow between consecutive frames using the Lucas-Kanade algorithm~\cite{lucas1981iterative}, then deducing long-term motions using forward flow accumulation. %
Since then, various approaches following the idea of chaining flows estimated between consecutive (or distant) frames have emerged~\cite{crivelli2014robust, crivelli2012optical, janai2017slow, wu2023accflow, neoral2023mft}.
Among them, AccFlow~\cite{wu2023accflow} introduces a backward accumulation technique, more robust to occlusions than forward strategies.
MFT~\cite{neoral2023mft} identifies the most reliable chain of optical flows by predicting uncertainty and occlusion scores.
Iterative methods, such as RAFT and its variants~\cite{teed2020raft, jiang2021learning, sui2022craft}, can be employed to warm-start the estimation of optical flow between distant frames. 
Recently, OmniMotion~\cite{wang2023omnimotion}, an optimization technique that relies on a volumetric representation, akin to a dynamic neural radiance field~\cite{mildenhall2021nerf}, have been proposed to estimate motion across every pixel and every frame in a video.
Similar ideas have also been applied to tracking points in 3D in a multi-camera setting~\cite{luiten2023dynamic}.

{\em \method{} uses a different strategy, relying on off-the-shelf point tracking algorithms to predict dense and long-term motions}, much faster than the per-video optimization technique in OmniMotion~\cite{wang2023omnimotion} and less prone to error accumulation than chaining methods like AccFlow~\cite{wu2023accflow} or MFT~\cite{neoral2023mft}.

\paragraph{Point tracking} is closely related to optical flow. Given a pixel in a frame, it predicts the position and visibility of the corresponding world surface point in every other frame of the video. 
This task has gained significant attention recently, particularly with the introduction of the TAP benchmark and the associated TAPNet baseline~\cite{doersch2022tap}. %
Numerous methods have since emerged in this domain.
Persistent independent particles (PIPs)~\cite{harley2022particle} is a method that tracks points even under occlusion, drawing inspiration from the concept of particle video~\cite{sand2008particle}, which lies midway between optical flow and local feature matching.
TAPIR~\cite{doersch2023tapir} combines the global matching strategy of TAPNet with the refinement step of PIPs. 
PIPs++~\cite{zheng2023pointodyssey} significantly extends the temporal window of PIPs, enhancing its robustness to long occlusion events. 
In contrast to previous methods which track one query at a time, CoTracker~\cite{karaev2023cotracker} uses a few additional tracks as context, resulting in improved performance. 

{\em \method{} takes this concept a step further by predicting the motion of all points in a frame simultaneously}, making it significantly more efficient (at least $\times$100 speedup) than point tracking methods applied densely, while matching or surpassing their performance on individual queries.
\section{Method}
\label{sec:method}

\paragraph{Overview.} We consider a sequence of RGB frames $X_t$ ($t$ in $1 \ldots T$) in $\RR^{W \times H \times 3}$. Given source and target time steps $(s,t)$, our goal is to predict for each pixel position $(x,y)$ in the source frame $X_s$ the visibility $v$ ($0$ if occluded and $1$ if visible) and the 2D location $(x+\Delta{x},y+\Delta{y})$ of the corresponding physical point in the target frame $X_t$. We represent these \emph{dense} correspondences between source and target as a flow $F_{{s}\rightarrow{t}}$ in $\RR^{W \times H \times 2}$ where $F_{{s}\rightarrow{t}}(x,y)=(\Delta{x},\Delta{y})$ and a mask $M_{{s}\rightarrow{t}}$ in $\{0, 1\}^{W \times H}$ where $M_{{s}\rightarrow{t}}(x,y)=v$.

\begin{figure}
    \centering
    \includegraphics[width=0.9\linewidth]{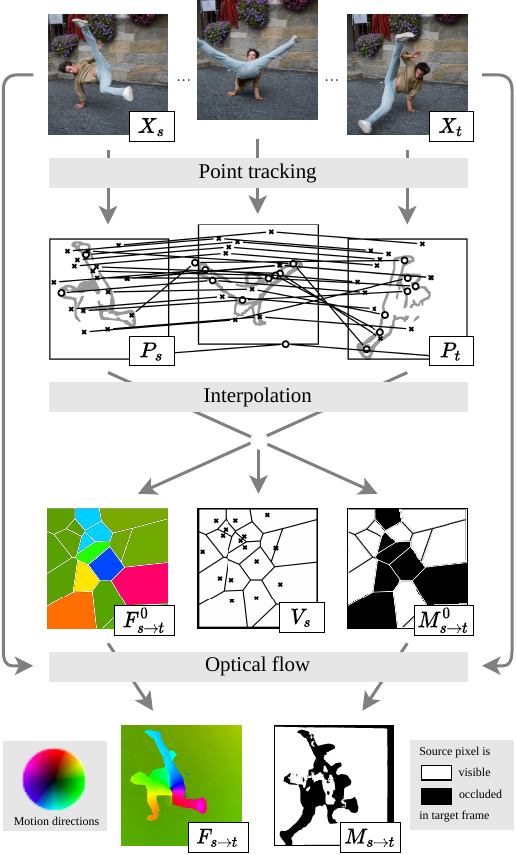}
    \caption{\textbf{Dense optical tracking.} Our approach, \method{}, takes a video as input and produces dense motion information between any pair of source and target frames $X_s$ / $X_t$ as an optical flow map $F_{{s}\rightarrow{t}}$ and a visibility mask $M_{{s}\rightarrow{t}}$. We first track the 2D position and the visibility (\setBold[0.5]$\times$\unsetBold: visible, \raisebox{-0.05\normalbaselineskip}[0pt][0pt]{\large\setBold[0.5]$\circ$\unsetBold}: occluded) of a small set of physical points throughout the video. These are sampled preferably from key regions at motion boundaries (shown in grey). 
    We deduce motion estimates $F^0_{{s}\rightarrow{t}}$ / $M^0_{{s}\rightarrow{t}}$ by using all the tracks whose associated point is visible at $s$, noted $V_s$, to initialize their nearest neighbors, forming Voronoi cells. 
    We finally refine these estimates with optical flow techniques, using the frames $X_s$ and $X_t$.} %
    \label{fig:densification}
\end{figure}

Our approach to \textit{dense optical tracking} (\method{}), illustrated in Figure~\ref{fig:densification}, is composed of the following modules: %
\begin{itemize}
    \item \textbf{Point tracking.} To accommodate long-range motions, we first compute $N$ tracks using an off-the-shelf point tracking method~\cite{karaev2023cotracker, doersch2023tapir, zheng2023pointodyssey} on all frames of the video. The number $N$ is kept low (${\sim}10^3$) compared to the number $HW$ of pixels (${\sim}10^5$) to limit computation and allow fast inference. A point $i$ in $1 \ldots N$ tracked at time $t$ in $1 \ldots T$ is denoted by $p^i_t=(x^i_t,y^i_t,v^i_t)$, with position $(x^i_t,y^i_t)$ in $\RR^2$, and visibility $v^i_t$ in $\{0, 1\}$. We also denote by $P_t=\{p^i_t \,\, | \,\, i \in [1\ldots N]\}$ the set with all the points at $t$. %

    \item \textbf{Interpolation.} Tracks offer \emph{sparse} correspondences between $s$ and $t$. We deduce initial flow and mask estimates, $F^0_{{s}\rightarrow{t}}$ and $M^0_{{s}\rightarrow{t}}$, using nearest-neighbor interpolation: we associate with every pixel in the source frame $X_s$ the nearest track $i$ among those visible at $s$, noted $V_s$, and use the position and visibility of the correspondence $(p^i_s,p^i_t)$ to initialize the flow and the mask respectively. %
    
    \item \textbf{Optical flow.} We refine these estimates into final predictions, $F_{{s}\rightarrow{t}}$ and $M_{{s}\rightarrow{t}}$, with an optical flow method inspired by RAFT~\cite{teed2020raft} which uses the source and target frames, $X_s$ and $X_t$, to account for the local geometry of objects. We obtain dense tracks by considering all frames as target for a given source, \eg, setting $s$ to $1$ and using $t$ in $\{2 \ldots T \}$ to get all tracks from the first frame. %

\end{itemize}
These components and the corresponding training processes are described in more detail in the following paragraphs.

\subsection{Overall approach} %
\label{sec:overall}

\textbf{Point tracking.} Not all tracks are equally informative and point tracking is computationally intensive, so we sample tracks more densely in regions undergoing significant motions or likely to become occluded. These are often localized around the edges of moving objects, at the boundary between pixels with significant motions and others which are either just dis-occluded or soon to be occluded. We find these regions by running a pre-trained optical flow model on consecutive frames of the video and then applying a Sobel filter~\cite{kanopoulos1988design} to detect discontinuities in the flow. %

Given a budget of $N$ tracks, our sampling strategy consists in initializing half of the tracks randomly near flow edges (up to $5$ pixels from an edge) and sampling the remaining ones in the entire image. 
It is important to emphasize that our approach is agnostic to the choice of a specific point tracking method,
so it will also benefit from ongoing advancements in this rapidly evolving research field.

\paragraph{Interpolation.} We initialize coarse motion and visibility estimates, $F^0_{{s}\rightarrow{t}}$ and $M^0_{{s}\rightarrow{t}}$, between source and target at a reduced spatial resolution $W/P \stimes H/P$ where $P$ is some constant ($P{=}4$ in practice). %
These are derived from the $N$ input tracks using the nearest visible track for every position $(x,y)$ in $[P, 2P, .., W] \times [P, 2P, .., H]$ in the source frame:%
\begin{gather}
    \left\{\begin{array}{ll}
    F^0_{{s}\rightarrow{t}}(x,y)&=(x^i_t,y^i_t) - (x^i_s,y^i_s), \\
    M^0_{{s}\rightarrow{t}}(x,y)&=v_t^i,
    \end{array} \right.
\end{gather}
with $i=\argmin_{j\in V_s}||(x^j_s,y^j_s) - (x,y)||_2$ and
where $V_s=\{j \in 1..N \,\, | \,\, v_j^s=1\}$ designates all the points which are visible at time $s$. Nearest neighbors of these points form Voronoi cells in the image plane, see Figure~\ref{fig:densification}.

\paragraph{Optical flow.} The estimates $F^0_{{s}\rightarrow{t}}$ and $M^0_{{s}\rightarrow{t}}$ are inherently imprecise, particularly for points distant from the input tracks.
We refine them using an optical flow method.

As in RAFT~\cite{teed2020raft}, we extract coarse features $Y_s$ (resp. $Y_t$) in $\RR^{W/P \times H/P \times C}$ for the source (resp. target) frame $X_s$ (resp. $X_t$) using a convolutional neural network.
We then compute the correlation between features for all pairs of source and target positions at different feature resolution levels, yielding a 4D correlation volume for each resolution level.
The flow is progressively refined by repeating $K$ times the following procedure ($K{=}4$ in practice):
Let $F^k_{{s}\rightarrow{t}}$ be the estimate at the iteration $k$. %
For each source position, we sample the correlation volume at different target positions laid on a regular grid centered at the position indicated by $F^{k}_{{s}\rightarrow{t}}$.
The similarity of each source point with a local neighborhood around its current target correspondence is then fed to a recurrent neural network~\cite{cho2014properties} to predict $F^{k+1}_{{s}\rightarrow{t}}$.
We obtain the flow $\widehat{F}_{{s}\rightarrow{t}}$ from the final estimate $F^K_{{s}\rightarrow{t}}$ using the upsampling operation introduced in RAFT~\cite{teed2020raft}.

Our model differs from RAFT~\cite{teed2020raft} in that we have a meaningful initialization for the flow $F^0_{{s}\rightarrow{t}}$ instead of a zero motion.
Moreover, we handle occlusions by refining a mask $\widehat{M}_{{s}\rightarrow{t}}$ from the coarse estimate $M^0_{{s}\rightarrow{t}}$ along with the flow. We note that the predicted mask $\widehat{M}_{{s}\rightarrow{t}}$ is a soft estimate which may be turned into a binary one by thresholding with a fixed scalar $\tau$ for every position ($\tau{=}0.8$ in practice).%

\subsection{Training process}

We use an off-the-shelf model for point tracking (\eg,~\cite{karaev2023cotracker, doersch2023tapir, zheng2023pointodyssey}) and freeze the corresponding parameters during training.
The interpolation is a parameter-free operation.
Therefore, among the three components of our approach, only the optical flow module requires training.

\paragraph{Objectives.} The parameters of \method{} are optimized by minimizing the sum of two objective functions: %
a motion reconstruction objective which is the $L_1$ distance between the predicted and the ground-truth flows, and a visibility prediction objective which is the binary cross entropy between the predicted and the ground-truth masks.

\begin{table*}
\setlength\heavyrulewidth{.25ex}
\aboverulesep=0ex
\belowrulesep=.3ex
\centering
\caption{\textbf{Motion prediction on the CVO benchmark.} We evaluate dense predictions between the first and last frames of videos. We report the end point error (EPE) of flows for all pixels (all), visible (vis) and occluded ones (occ), the intersection over union (IoU) of occluded regions, and the average inference time per video (in seconds). We also indicate the number $N$ of tracks for different methods.} %
\vspace{-.2em}
\small

\begin{tabular}{@{}L{1.1cm}@{}L{2.3cm}@{}C{1.3cm}@{}@{}C{2.8cm}@{}C{0.75cm}@{}C{2.8cm}@{}C{0.75cm}@{}C{1.0cm}@{}C{2.8cm}@{}C{0.75cm}@{}C{1.05cm}@{}} %
\toprule
\multicolumn{2}{@{}l@{}}{\multirow{2}{*}{\normalsize Method}} & \multirow{2}{*}{$N$} & \multicolumn{2}{@{}c@{}}{CVO (\textit{Clean})} & \multicolumn{3}{@{}c@{}}{CVO (\textit{Final})} & \multicolumn{3}{@{}c@{}}{CVO (\textit{Extended})} \\ %
\cmidrule(l){4-5}\cmidrule(l){6-8}\cmidrule(l){9-11}
& & & EPE $\downarrow$ ({\scriptsize all / vis / occ}) & IoU $\uparrow$ & EPE $\downarrow$ ({\scriptsize all / vis / occ}) & IoU  $\uparrow$ & Time$^*$$\downarrow$ & EPE $\downarrow$ ({\scriptsize all / vis / occ}) & IoU $\uparrow$ & Time $\downarrow$ \\
\midrule
\multirow{6}{*}{~~~\raisebox{0.05\normalbaselineskip}[0pt][0pt]{\rotatebox[origin=c]{90}{Optical flow}}} & RAFT~\cite{teed2020raft} & - & 2.82 / 1.70 / 8.01 & 58.1 & 2.88 / 1.79 / 7.89 & 57.2 & \textbf{0.166} & 28.6 / 21.6 / 41.0 & 61.7 & \textbf{0.166} \\
& GMA~\cite{jiang2021learning} & - & 2.90 / 1.91 / 7.63 & 60.9 & 2.92 / 1.89 / 7.48 & 60.1 & \underline{0.186} & 30.0 / 22.8 / 42.6 & 61.5 & \underline{0.186} \\
& RAFT (\raisebox{-0.05em}{\includegraphics[height=0.65em]{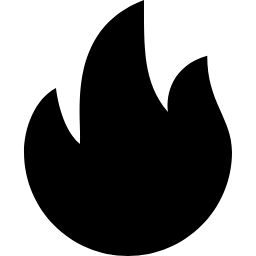}})~\cite{teed2020raft} & - & 2.48 / 1.40 / 7.42 & 57.6 & 2.63 / 1.57 / 7.50 & 56.7 & 0.634 & 21.8 / 15.4 / 33.4 & 65.0 & 4.142 \\
& GMA (\raisebox{-0.05em}{\includegraphics[height=0.65em]{files/figures/flame.png}})~\cite{jiang2021learning} & - & 2.42 / 1.38 / 7.14 & 60.5 & 2.57 / 1.52 / 7.22 & 59.7 & 0.708 & 21.8 / 15.7 / 32.8 & 65.6 & 4.796 \\
& MFT~\cite{neoral2023mft} & - & 2.91 / 1.39 / 9.93 & 19.4 & 3.16 / 1.56 / 10.3 & 19.5 & 1.350 & 21.4 / 9.20 / 41.8 & 37.6 & 18.69\\
& AccFlow~\cite{wu2023accflow} & - & 1.69 / 1.08 / 4.70 & 48.1 & 1.73 / 1.15 / 4.63 & 47.5 & 0.746 & 36.7 / 28.1 / 52.9 & 36.5 & 5.598 \\
\midrule
\\[-1em]
\multirow{3}{*}{\raisebox{0.05\normalbaselineskip}[0pt][0pt]{\rotatebox[origin=c]{90}{\begin{tabular}{@{}c@{}} Point \\ tracking\end{tabular}}}} & PIPs++~\cite{zheng2023pointodyssey} & 262144 & 9.05 / 6.62 / 21.5 & 33.3 & 9.49 / 7.06 / 22.0 & 32.7 & 974.3 & 18.4 / 10.0 / 32.1 & 58.7 & 1922. \\
& TAPIR~\cite{doersch2023tapir} & 262144 & 3.80 / 1.49 / 14.7 & 73.5 & 4.19 / 1.86 / 15.3 & 72.4 & 131.1 & 19.8 / 4.74 / 42.5 & 68.4 & 848.7 \\ 
& CoTracker~\cite{karaev2023cotracker} & 262144 & {1.51} / {0.88} / {4.57} & {75.5} & {1.52} / {0.93} / {4.38} & {75.3} & 173.5 & {5.20} / 3.84 / {7.70} & {70.4} & 1645. \\[0.25em]
\midrule
\\[-1em]
\multirow{3}{*}{~~~\raisebox{0.05\normalbaselineskip}[0pt][0pt]{\rotatebox[origin=c]{90}{Hybrid}}} & %
\multirow{3}{*}{\begin{tabular}{@{}l@{}} \textit{Dense optical} \\ \textit{tracking} (\method{})\end{tabular}}
& 1024 & {1.36} / {0.76} / {4.26} & {80.0} & {1.43} / {0.85} / {4.29} & {79.7} & 0.864 & {5.28} / {3.78} / {7.71} & {70.8} & 5.234 \\
& & 2048 & \underline{1.32} / \underline{0.74} / \underline{4.12} & \underline{80.4} & \underline{1.38} / \underline{0.82} / \underline{4.10} & \underline{80.2} & 1.652 & \underline{5.07} / \underline{3.67} / \underline{7.34} & \underline{71.0} & 9.860 \\
& & 4096 & \textbf{1.29} / \textbf{0.72} / \textbf{4.03} & \textbf{80.4} & \textbf{1.34} / \textbf{0.80} / \textbf{3.99} & \textbf{80.4} & 3.152 & \textbf{4.98} / \textbf{3.59} / \textbf{7.17} & \textbf{71.1} & 19.73 \\[0.25em]

\bottomrule 
\end{tabular}
{\scriptsize ``$*$'': the time is the same for \textit{Clean} and \textit{Final} sets.}
\vspace{-1.1em}

\label{tab:cvo_quant}
\end{table*}

\newcommand{\imageWithText}[3]{%
\begin{tikzpicture}
    \node[anchor=south west, inner sep=0, outer sep=0] (image) at (0,0) {\includegraphics[width=#3]{#1}};
    
    \node[anchor=south west, text=black] at (0,0) {\setlength{\fboxsep}{2pt} \footnotesize \colorbox{white}{#2}};
\end{tikzpicture}
\hspace{-0.5em}
}

\begin{figure*}
    \newcommand{\myTime}[1]{\setlength{\fboxsep}{2pt} \footnotesize \colorbox{gray}{\textcolor{white}{#1}}}
	\setlength\tabcolsep{1.pt}
	\renewcommand{\arraystretch}{0.5}
	\small
    \centering
    \setlength{\fboxsep}{0pt}
    \setlength{\fboxrule}{1pt} 
	\begin{tabular}{@{}c@{}c@{}ccc@{}c@{}c@{}c@{}c@{}c@{}c@{}}

         \imageWithText{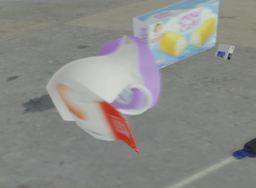}{Source}{.162\linewidth} & ~ & \imageWithText{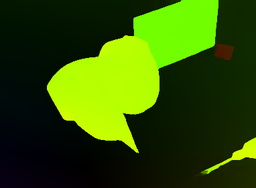}{RAFT}{.162\linewidth} &
        \imageWithText{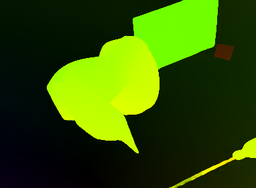}{GMA}{.162\linewidth} &
        \imageWithText{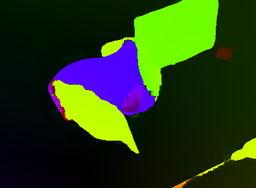}{AccFlow}{.162\linewidth} & ~ & 
        \imageWithText{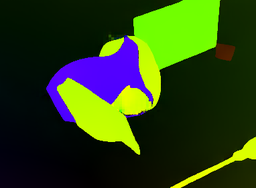}{\method{}}{.162\linewidth} & ~ &
        \hspace{-0.25em}\imageWithText{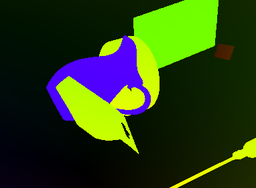}{Ground truth}{.162\linewidth} \\ [-6.65em]

        & & \multicolumn{3}{@{}c@{}}{\setlength{\fboxsep}{1pt}\colorbox{mygray}{\raisebox{0.15em}{\textcolor{black}{~~~~~~~~~~~~~~~~~~~~~~~~~~~~~~~~~~~~~~~~~~~~~{\footnotesize Optical Flow}~~~~~~~~~~~~~~~~~~~~~~~~~~~~~~~~~~~~~~~~~~~}}}} & & \multicolumn{1}{@{}c@{}}{\setlength{\fboxsep}{1pt}\colorbox{mygray}{\raisebox{0.15em}{\textcolor{black}{~~~~~~~~~~~~{\footnotesize Hybrid}~~~~~~~~~~~}}}}\\[5.65em]

         \imageWithText{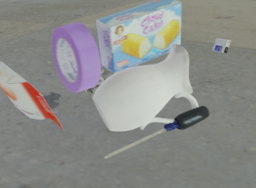}{Target}{.162\linewidth} & & \raisebox{0.1em}[0pt][0pt]{\fbox{\includegraphics[width=.158\linewidth]{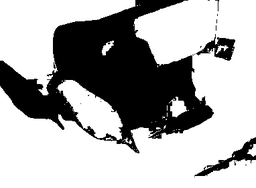}}} &
        \raisebox{0.1em}[0pt][0pt]{\fbox{\includegraphics[width=.158\linewidth]{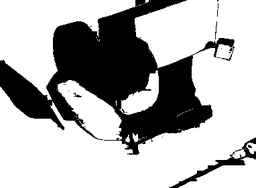}}} &
        \raisebox{0.1em}[0pt][0pt]{\fbox{\includegraphics[width=.158\linewidth]{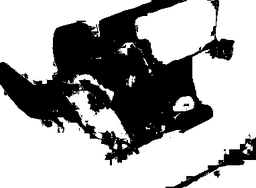}}} & &
        \raisebox{0.1em}[0pt][0pt]{\fbox{\includegraphics[width=.158\linewidth]{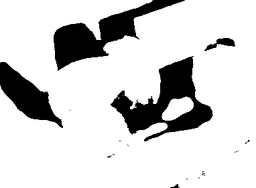}}} & &
        \raisebox{0.1em}[0pt][0pt]{\fbox{\includegraphics[width=.158\linewidth]{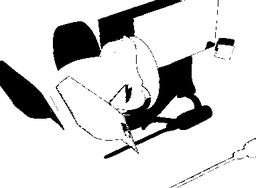}}} \\[-1.5em]
        & & \myTime{\footnotesize 166 milliseconds} & \myTime{\footnotesize 186 milliseconds} & \myTime{\footnotesize 746 milliseconds} & & \myTime{\footnotesize 864 milliseconds} \\ [0.8em]

        \imageWithText{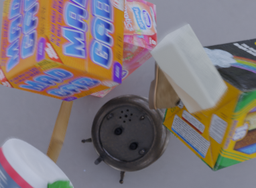}{Source}{.162\linewidth} & &\imageWithText{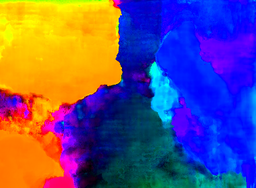}{PIPs++}{.162\linewidth} &
        \imageWithText{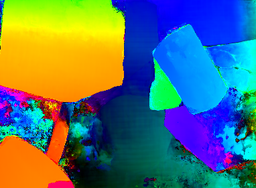}{TAPIR}{.162\linewidth} &
        \imageWithText{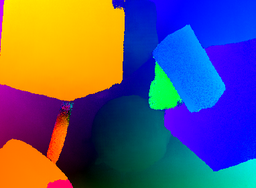}{CoTracker}{.162\linewidth} & &
        \imageWithText{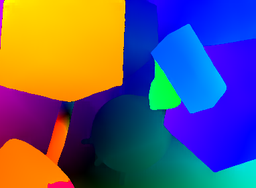}{\method{}}{.162\linewidth} & &
        \hspace{-0.25em}\imageWithText{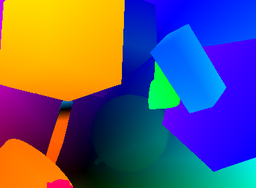}{Ground truth}{.162\linewidth} \\ [-6.65em]

        & & \multicolumn{3}{@{}c@{}}{\setlength{\fboxsep}{1pt}\colorbox{mygray}{\raisebox{0.15em}{\textcolor{black}{~~~~~~~~~~~~~~~~~~~~~~~~~~~~~~~~~~~~~~~~~~~~{\footnotesize Point tracking}~~~~~~~~~~~~~~~~~~~~~~~~~~~~~~~~~~~~~~~~~~}}}} & & \multicolumn{1}{@{}c@{}}{\setlength{\fboxsep}{1pt}\colorbox{mygray}{\raisebox{0.15em}{\textcolor{black}{~~~~~~~~~~~{\footnotesize Hybrid}~~~~~~~~~~~~}}}}\\[5.65em]

        \imageWithText{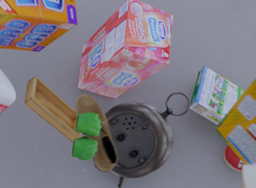}{Target}{.162\linewidth} & & \raisebox{0.1em}[0pt][0pt]{\fbox{\includegraphics[width=.158\linewidth]{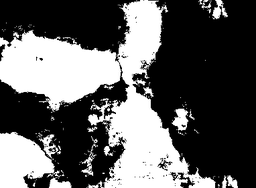}}} &
        \raisebox{0.1em}[0pt][0pt]{\fbox{\includegraphics[width=.158\linewidth]{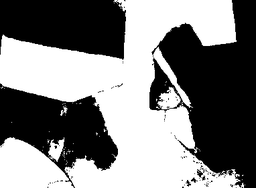}}} &
        \raisebox{0.1em}[0pt][0pt]{\fbox{\includegraphics[width=.158\linewidth]{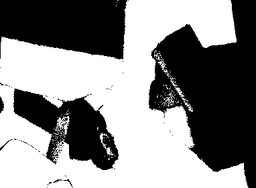}}} & &
        \raisebox{0.1em}[0pt][0pt]{\fbox{\includegraphics[width=.158\linewidth]{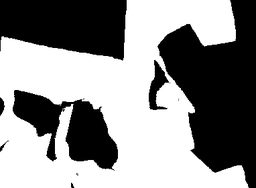}}} & &
        \raisebox{0.1em}[0pt][0pt]{\fbox{\includegraphics[width=.158\linewidth]{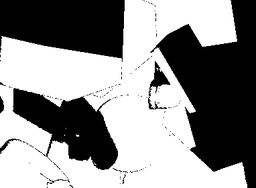}}} \\[-1.5em]
        & & \myTime{\footnotesize 16 minutes} & \myTime{\footnotesize 2 minutes} & \myTime{\footnotesize 3 minutes} & & \myTime{\footnotesize 864 milliseconds} \\[-0.3em]
		
	\end{tabular}
	\caption{\textbf{Qualitative samples on the CVO benchmark.} We show the predicted flow (1st / 3rd rows) and visibility mask (2nd / 4th rows) between the first and last frame of different videos of the \textit{Final} test set. We also report inference time. Optical flow methods produce smooth motion estimates but miss important object regions. Point tracking methods, PIPS++ and TAPIR, are more accurate but tend to produce noisy estimates. CoTracker improves on this aspect by processing multiple point tracks simultaneously instead of one at a time, but we still observe some artifacts when zooming in. \method{} combines the benefits of both optical flow and point tracking approaches.\cs{figure too overload, very difficult to understand, even if it goes into the exp. we need to give more details and guide the reader}}
    \vspace{-0.2em}
	\label{fig:cvo_qual}
\end{figure*}

\input{files/tables/cvo_speed}

\paragraph{Optimization.} \method{} is trained on frames at res.\ $512 \times 512$ for 500k steps with the ADAM optimizer~\cite{kingma2014adam} and a learning rate of $10^{-4}$ using $4$ NVIDIA V100 GPUs. Given the practical challenge of gathering and storing dense ground truth across different time horizons, \eg, a single flow map represents 262,144 correspondences at this resolution, we only compute the training objectives on a few of these correspondences.
Specifically, we train using $N{=}2048$ tracks from CoTracker~\cite{karaev2023cotracker} as input (see ablations with other methods in Table~\ref{tab:cvo_corr_abl}), and use another $N$ ground-truth tracks (synthetically generated) for supervision.
At inference, we adjust $N$ to trade motion prediction quality for speed (Table~\ref{fig:cvo_speed}). %
Further implementation details are presented in Appendix~\ref{sec:arch}-\ref{sec:inter}.
For reproducibility, code, data and pretrained models are publicly available in our project webpage.

\cs{which dataset? synthetic data?}

\section{Experiments}
\label{sec:experiments}

\begin{table*}
\setlength\heavyrulewidth{.25ex}
\aboverulesep=0ex
\belowrulesep=.3ex
\centering
\caption{\textbf{Motion prediction on the TAP benchmark.} We evaluate trajectories spanning the whole video for specific points. We report average jaccard (AJ), position accuracy ({\scriptsize$<$}$\delta^x_\text{avg}$), and occlusion accuracy (OA). For completeness we include single-point methods optimized for individual test queries, but report their performance separately since they are not strictly comparable when evaluating dense predictions. }
\vspace{-.2em}
\small

\begin{tabular}{@{}L{0.6cm}@{}L{2.55cm}@{}C{0.95cm}@{}@{}C{0.95cm}@{}C{0.95cm}@{}C{0.95cm}@{}C{0.95cm}@{}C{0.95cm}@{}C{0.95cm}@{}C{0.95cm}@{}C{0.95cm}@{}C{0.95cm}@{}C{0.95cm}@{}C{0.95cm}@{}C{0.95cm}@{}C{0.95cm}@{}C{0.95cm}@{}} %
\toprule
\multicolumn{2}{@{}l@{}}{\multirow{2}{*}{\normalsize Method}} & \multicolumn{3}{@{}c@{}}{DAVIS (\textit{First})} & \multicolumn{3}{@{}c@{}}{DAVIS (\textit{Strided})} & \multicolumn{3}{@{}c@{}}{RGB-S. (\textit{First})} & \multicolumn{3}{@{}c@{}}{RGB-S. (\textit{Strided})} & \multicolumn{3}{@{}c@{}}{Kinetics (\textit{First})} \\
\cmidrule(l){3-5}\cmidrule(l){6-8}\cmidrule(l){9-11}\cmidrule(l){12-14}\cmidrule(l){15-17}
& & AJ $\uparrow$ & {\scriptsize$<$}$\delta^x_\text{avg}$ $\uparrow$ & OA $\uparrow$ & AJ $\uparrow$ & {\scriptsize$<$}$\delta^x_\text{avg}$ $\uparrow$ & OA $\uparrow$ & AJ $\uparrow$ & {\scriptsize$<$}$\delta^x_\text{avg}$ $\uparrow$ & OA $\uparrow$ & AJ $\uparrow$ & {\scriptsize$<$}$\delta^x_\text{avg}$ $\uparrow$ & OA $\uparrow$ & AJ $\uparrow$ & {\scriptsize$<$}$\delta^x_\text{avg}$ $\uparrow$ & OA $\uparrow$ \\
\midrule
\multirow{4}{*}{\raisebox{0.05\normalbaselineskip}[0pt][0pt]{\rotatebox[origin=c]{90}{Single}}} & \textcolor{black}{TAP-Net~\cite{doersch2022tap}} & \textcolor{black}{33.0} & \textcolor{black}{48.6} & \textcolor{black}{78.8} & \textcolor{black}{38.4} & \textcolor{black}{53.1} & \textcolor{black}{82.3} & \textcolor{black}{53.5} & \textcolor{black}{68.1} & \textcolor{black}{86.3} & \textcolor{black}{{59.9}} & \textcolor{black}{{72.8}} & \textcolor{black}{{90.4}} & \textcolor{black}{38.5} & \textcolor{black}{54.4} & \textcolor{black}{80.6} \\
& \textcolor{black}{PIPs~\cite{harley2022particle}} & \textcolor{black}{42.2} & \textcolor{black}{64.8} & \textcolor{black}{77.7} & \textcolor{black}{52.4} & \textcolor{black}{70.0} & \textcolor{black}{83.6} & \textcolor{black}{-} & \textcolor{black}{-} & \textcolor{black}{-} & \textcolor{black}{-} & \textcolor{black}{-} & \textcolor{black}{-} & \textcolor{black}{31.7} & \textcolor{black}{53.7} & \textcolor{black}{72.9} \\
& \textcolor{black}{TAPIR~\cite{doersch2023tapir}} & \textcolor{black}{\underline{56.2}} & \textcolor{black}{\underline{70.0}} & \textcolor{black}{\underline{86.5}} & \textcolor{black}{\underline{61.3}} & \textcolor{black}{\underline{73.6}} & \textcolor{black}{\textbf{88.8}} & \textcolor{black}{\underline{55.5}} & \textcolor{black}{\underline{69.7}} & \textcolor{black}{\underline{88.0}} & \textcolor{black}{\underline{62.7}} & \textcolor{black}{\underline{74.6}} & \textcolor{black}{\underline{91.6}} & \textcolor{black}{\textbf{49.6}} & \textcolor{black}{\underline{64.2}} & \textcolor{black}{\underline{85.0}} \\
& \textcolor{black}{CoTracker (\raisebox{-0.05em}{\large $\filledstar$})~\cite{karaev2023cotracker}} & \textcolor{black}{\textbf{60.6}} & \textcolor{black}{\textbf{75.4}} & \textcolor{black}{\textbf{89.3}} & \textcolor{black}{\textbf{64.8}} & \textcolor{black}{\textbf{79.1}} & \textcolor{black}{\underline{88.7}} & \textcolor{black}{\textbf{65.4}} & \textcolor{black}{\textbf{79.1}} & \textcolor{black}{\textbf{91.0}} & \textcolor{black}{\textbf{73.6}} & \textcolor{black}{\textbf{84.5}} & \textcolor{black}{\textbf{94.5}} & \textcolor{black}{\underline{48.7}} & \textcolor{black}{\textbf{64.3}} & \textcolor{black}{\textbf{86.5}} \\
\midrule
\\[-1em]
\midrule
\multirow{5}{*}{\raisebox{0.05\normalbaselineskip}[0pt][0pt]{\rotatebox[origin=c]{90}{Dense}}} & RAFT (\raisebox{-0.05em}{\includegraphics[height=0.65em]{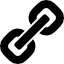}})~\cite{teed2020raft} & - & - & - & 30.0 & 46.3 & 79.6 & - & - & - & 44.0 & 58.6 & 90.4 & - & - & - \\
& OmniMotion~\cite{wang2023omnimotion} & - & - & - & 51.7 & 67.5 & 85.3 & - & - & - & \underline{77.5} & \underline{87.0} & \underline{93.5} & - & - & - \\
& MFT~\cite{neoral2023mft} & {47.3} & {66.8} & {77.8} & {56.1} & {70.8} & \underline{86.9} & - & - & - & - & - & - & 39.6 & 60.4 & 72.7 \\
& CoTracker~\cite{karaev2023cotracker} & \underline{56.9} & \underline{74.1} & \underline{84.4} & \underline{61.1} & \underline{77.4} & {85.8} & \underline{67.7} & \underline{80.7} & \underline{90.8} & {74.1} & {85.2} & {92.3} & \underline{44.8} & \underline{63.2} & \underline{81.2} \\
& \method{} (\textit{ours}) & \textbf{60.1} & \textbf{74.5} & \textbf{89.0} & \textbf{65.9} & \textbf{79.2} & \textbf{90.2} & \textbf{77.1} & \textbf{87.7} & \textbf{93.3} & \textbf{83.4} & \textbf{91.4} & \textbf{95.7} & \textbf{48.4} & \textbf{63.8} & \textbf{85.2} \\
\bottomrule 
\end{tabular}
\label{tab:tap_quant}
\end{table*}

\begin{figure*}
    \newcommand{\withText}[3]{%
    \begin{tikzpicture}
        \node[anchor=south west, inner sep=0, outer sep=0] (image) at (0,0) {#1};
        \node[anchor=south west, text=black] at (-0.12,1.45) {\setlength{\fboxsep}{2pt} \footnotesize \colorbox{white}{#2}};
        \node[anchor=south west, text=white] at (-0.12,1.12) {\setlength{\fboxsep}{2pt} \footnotesize \colorbox{gray}{#3}};
    \end{tikzpicture}
    \hspace{-0.5em}}
    \newcommand{\withName}[2]{%
    \begin{tikzpicture}
        \node[anchor=south west, inner sep=0, outer sep=0] (image) at (0,0) {#1};
        \node[anchor=south west, text=black] at (-0.12,1.45) {\setlength{\fboxsep}{2pt} \footnotesize \colorbox{white}{#2}};
    \end{tikzpicture}
    \hspace{-0.5em}}
    \newcommand{\hike}[1]{\includegraphics[width=0.163\linewidth]{#1}}
    \newcommand{\duckLarge}[1]{\includegraphics[width=0.163\linewidth]{#1}}
    \newcommand{\duck}[1]{\includegraphics[width=0.163\linewidth]{#1}}
    \newcommand{\duckright}[1]{\includegraphics[width=0.163\linewidth]{#1}}
    \newcommand{\drift}[1]{\includegraphics[width=0.163\linewidth]{#1}}

	\setlength\tabcolsep{1.pt}
	\renewcommand{\arraystretch}{0.25}
	\small
    \centering
    \setlength{\fboxsep}{0pt}
    \setlength{\fboxrule}{1pt}
    \hspace{-0.75em}
	\begin{tabular}{@{}cccccc@{}}
        \multicolumn{2}{@{}l@{}}{~~\raisebox{1.5em}{\begin{tabular}{@{}C{1.35cm}@{}}
        \textit{Hike}\\[0.25cm]{\scriptsize $s=1$}\\{\scriptsize $t\in\{25,50\}$}
        \end{tabular}}\withName{\hike{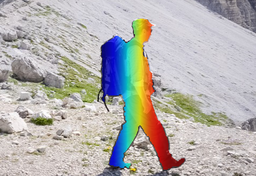}}{Source}} & 
        \multicolumn{2}{@{}l@{}}{\raisebox{1.5em}{\begin{tabular}{@{}C{1.5cm}@{}}
        \textit{Duck}\\[0.25cm]{\scriptsize $s=1$}\\{\scriptsize $t\in\{15,30\}$}
        \end{tabular}}\duck{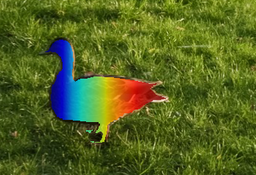}} &
        \multicolumn{2}{@{}l@{}}{\raisebox{1.5em}{\begin{tabular}{@{}C{1.5cm}@{}}
        \textit{Drift}\\[0.25cm]{\scriptsize $s=1$}\\{\scriptsize $t\in\{20,40\}$}
        \end{tabular}}\drift{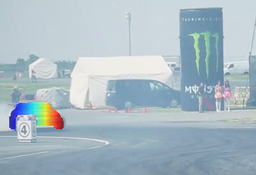}}\\
        
        \withText{\hike{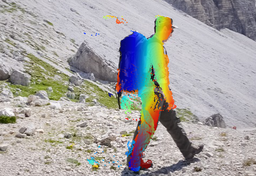}}{MFT}{1 minute}\hspace{0.25em} & \hike{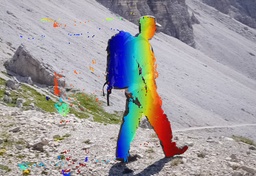} & \duck{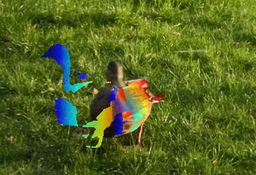} & \duckright{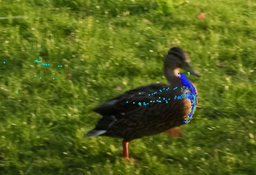} & \drift{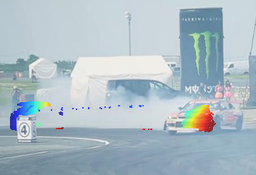} & \drift{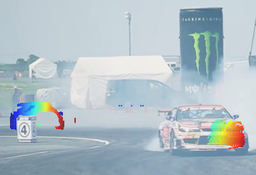}\\

        \withText{\hike{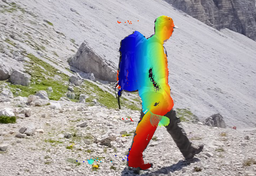}}{OmniMotion}{9 hours}\hspace{0.25em} & \hike{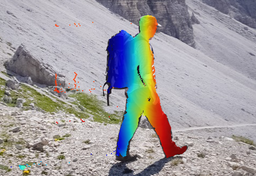} & \duck{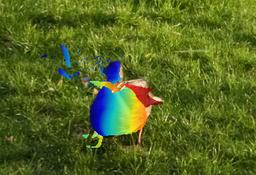} & \duckright{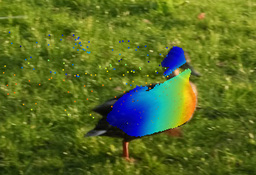} & \drift{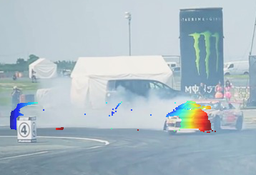} & \drift{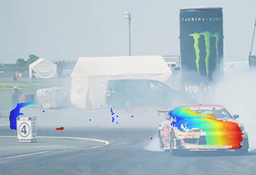} \\
        
        \withText{\hike{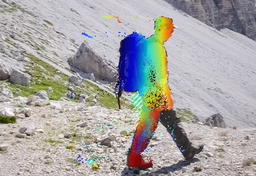}}{CoTracker}{1 hour}\hspace{0.25em} & \hike{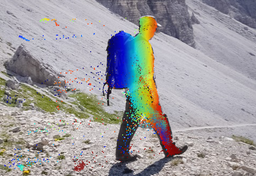} & \duck{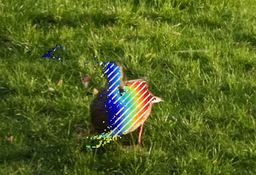} & \duckright{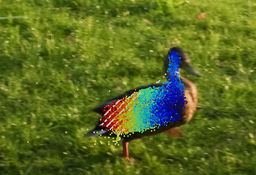} & \drift{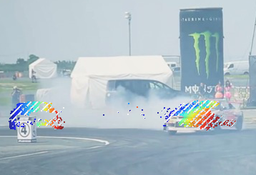} & \drift{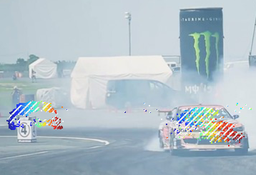} \\
        
        \withText{\hike{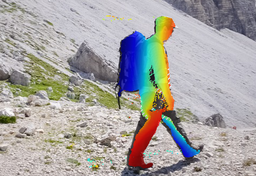}}{\method{}}{2 minutes}\hspace{0.25em} & \hike{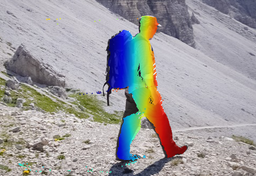} & \duck{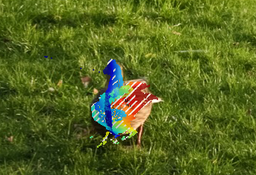} & \duckright{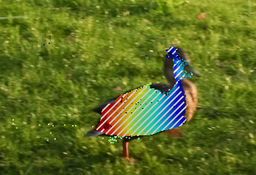} & \drift{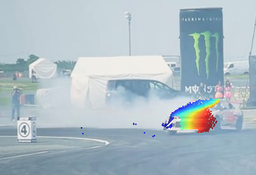} & \drift{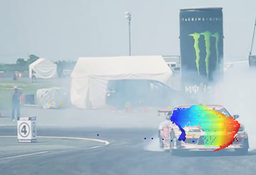}\\
		
	\end{tabular}
	\vspace{-0.3em}
	\caption{\textbf{Qualitative samples on the TAP benchmark.}  We compare various methods by tracking all points in the first frame of videos from the DAVIS dataset. Only foreground points are visualized, each depicted with distinctive colors, and overlayed with white stripes when occluded. We also indicate the time for each method to process a 480p video of 50 frames on an NVIDIA V100 GPU. In the ``\textit{Hike}'' video, our method, \method{}, stands out by successfully tracking both legs as the person walks. \method{} has robust performance under occlusion, as shown in the ``\textit{Duck}'' video where the animal changes sides. In contrast, MFT lose sight of the object, showing the limitations of optical flow methods under occlusion. OmniMotion does not account for the rotation of the object. CoTracker successfully tracks the object but fails to predict occlusions, showing the limitations of point tracking methods overly reliant on local features, especially when different parts of an object look similar. \method{} handles videos with small objects or atmospheric effects like smoke, like in the ``\textit{Drift}'' video. Other methods tend to miss object parts in similar conditions. Please zoom in for details and refer to the videos in the supplemental materials.}
	\label{fig:tap_qual}
\end{figure*}

\paragraph{Optical flow baselines} include RAFT~\cite{teed2020raft} and GMA~\cite{jiang2021learning}, methods which directly predict the motion between pairs of distant frames. We also employ these methods as {warm starts}, represented by (\raisebox{-0.05em}{\includegraphics[height=0.65em]{files/figures/flame.png}}) in our tables and figures, where the flow between neighboring frames serves as an initialization for estimating the flow between more distant frames. Another strategy, represented as (\raisebox{-0.05em}{\includegraphics[height=0.65em]{files/figures/chain.png}}), is to chain optical flows computed for adjacent frames. We also compare to AccFlow~\cite{wu2023accflow}, a backward accumulation method for estimating long-range motions, MFT~\cite{neoral2023mft}, an advanced method for chaining optical flows, and OmniMotion~\cite{wang2023omnimotion}, a method which regularizes optical flow by constructing a volumetric representation. All three methods build upon RAFT~\cite{teed2020raft}.

\paragraph{Point tracking baselines.} We explore the direct extension of sparse methods to predict dense motions by applying them at every pixel.
Specifically, we use PIPs~\cite{harley2022particle}, PIPs++~\cite{zheng2023pointodyssey}, TAP-Net~\cite{doersch2022tap} and TAPIR~\cite{doersch2023tapir}, all of which track individual point independently.
We also consider CoTracker~\cite{karaev2023cotracker}, which enables the simultaneous tracking of multiple points in two different modes: 
The first processes batches of points spread over the image and is hence efficient. 
The second one, which we denote as (\raisebox{-0.12em}{\Large $\filledstar$}), is specifically optimized for individual point queries, by incorporating a context of local and global points solely for inference, discarding them afterwards. 
While the latter improves precision, it does so at the expense of considerably slower processing, making it impractical for dense prediction.
Karaev \etal use the first version for visualizations and the second for quantitative evaluations~\cite{karaev2023cotracker}.
Given their significant differences, we treat them as distinct approaches.

\paragraph{Datasets.} Like others, we train and evaluate our method using data generated with {Kubric}~\cite{greff2022kubric}, a simulator for realistic rendering of RGB frames along with motion information, featuring scenes with objects falling to the ground and colliding with one another. We consider different datasets:
\begin{itemize}
\item \textbf{MOVi-F Train.} This set is composed of approximately 10,000 videos, each containing 24 frames rendered at 12 frames per second (FPS). The videos are equipped with point tracks, primarily sampled from objects and, to a lesser extent, from the background. Our method and point tracking baselines~\cite{doersch2023tapir, karaev2023cotracker, harley2022particle, doersch2022tap} train on this data, or close variations thereof. We note that PIPS++~\cite{zheng2023pointodyssey} uses data from a different simulator called PointOdyssey, focusing on long videos with naturalistic motion.
\item \textbf{CVO Train~\cite{wu2023accflow}.} This set has around 10,000 videos, with 7 frames rendered at 60 FPS. The videos come with bidirectional optical flow information for adjacent frames and between the first frame and every other frame. Our optical flow baselines~\cite{wu2023accflow, teed2020raft, jiang2021learning} use this data for training.
\item \textbf{CVO Test~\cite{wu2023accflow}.} There were originally two test sets, \textit{Clean} and \textit{Final}, with the latter incorporating motion blur. Each of them contains around 500 videos of 7 frames at 60 FPS.\footnote{We use a curated version of this dataset since in its original release it contained a few scenes with erroneous optical flows (for 25 videos). Our full data curation pipeline is detailed in Appendix~\ref{sec:cvo_curation}.} We also introduce the \textit{Extended} CVO set, with another 500 videos of 48 frames rendered at 24 FPS. This new set is designed to assess longer videos with more challenging motions. All sets provide the optical flow and visibility mask between the first and last frame of videos. %
\end{itemize}
~

\noindent We further evaluate \method{} and other approaches on the TAP benchmark~\cite{doersch2022tap}. It provides ground-truth tracks for various types of scenes, including real-world videos:
\textbf{DAVIS~\cite{perazzi2016benchmark},} with 30 videos ($\sim$100 frames each) featuring one salient object; %
\textbf{Kinetics~\cite{carreira2017quo},} with over 1,000 videos (250 frames each) representing various human actions;
\textbf{RGB-Stacking~\cite{lee2021beyond},} with 50 synthetic videos (250 frames each) in a robotic environment with textureless objects and frequent occlusions.

\paragraph{Evaluation metrics.} %
We measure computational efficiency and the quality of dense motion predictions between the first and last frames of videos using the following metrics: %
\begin{itemize}
    \item {End point error} (EPE)  between predicted and ground-truth flows. We give mean EPE for all pixels, as well as separately for visible (vis) and occluded (occ) pixels.
    \item {Intersection over union} (IoU) between predicted and ground-truth occluded regions in visibility masks.
    \item {Computational efficiency} as the average inference time required to produce the mask and flow between the first and last frame of one video using an NVIDIA V100 GPU. 
\end{itemize}
Some methods may exclusively produce flow predictions so we estimate the visibility mask by doing forward-backward consistency checks on the predicted flow. This involves processing videos a second time by flipping their temporal axis.

We follow the standard evaluation protocol for TAP~\cite{doersch2022tap} and report occlusion accuracy (OA), position accuracy for visible points averaged over different threshold distances ({\scriptsize$<$}$\delta^x_\text{avg}$), and average jaccard (AJ) which combines both.%

\subsection{Comparison with the state of the art}

\cs{does 262144 correspond to number of tracks you obtain with our 
method? please clarify, can we compare to OmniMotion?}

\paragraph{CVO.} Results in terms of EPE and IoU on the CVO test sets in Table~\ref{tab:cvo_quant} show that \method{} significantly improves over optical flow baselines. In particular, on the \textit{Extended} set, with large motions and long occlusion events, \method{} reduces by a factor 4 the EPE and yields relative improvements of more than 8\% in IoU compared to these methods.
\method{} also outperforms point tracking baselines while being at least two orders of magnitude faster.
These methods are slow, even when parallelizing computations on GPU, as they are applied to every of the $N=512 \times 512=262144$ pixels in a frame.
\method{}, in contrast, is flexible as it may trade motion prediction quality for speed by adjusting the number $N$ of initial point tracks.
See also Appendix~\ref{sec:tap_tracks} for visual comparisons by taking different values for $N$.
Figure~\ref{fig:cvo_speed} shows that \method{} yields the best possible trade-offs.
\camera{DOT even improves over input tracks. By extracting $1024$ tracks with CoTracker, we obtain 10.6\%
(resp.\ 7.1\%) relative improvements on EPE (resp.\ IoU) on the \textit{Final} set on the same tracks after refining them with DOT.}
Qualitative samples in Figure~\ref{fig:cvo_qual} show the superiority of \method{} over prior works.

\paragraph{TAP.} \method{} predicts dense motions, without knowing which points will be used for testing, as opposed to single-point tracking techniques, optimized for specific test queries.
Remarkably, even under this challenging setting, \method{} is competitive with the best-performing single-point tracking algorithms across all of the TAP benchmark test sets, see Table~\ref{tab:tap_quant}.
\method{} even slightly improves over the state of the art on DAVIS (\textit{strided}), and achieves a substantial advantage on RGB-Stacking, with over 13\% relative increase in average jaccard (AJ) compared to single-point methods.
Its success lies in its ability to handle textureless objects effectively, whose lack of distinctive local features make them challenging for point tracking approaches.
The optical flow component in our approach allows us to rely on a broader context, thereby significantly enhancing motion predictions for such objects.
Furthermore, \method{} performs significantly better than dense methods which, like our method, are not optimized for specific test queries. The relative improvements range from 6\% to 15\% in average jaccard (AJ), up to 9\% in position accuracy ($\delta^x_\text{avg}$) and up to 5\% in occlusion accuracy (OA) compared to the previous state of the art.
Qualitative results on real data in Figures~\ref{fig:tap_qual}-\ref{fig:tap_struct}, \camera{in Appendix~\ref{sec:tap_space_time} and~\ref{sec:tap_changes}}, and videos in our project webpage show the significant improvements achieved by our method in terms of spatial consistency, robustness to occlusions and \camera{appearance changes} compared to the state of the art.

\begin{table}
\setlength\heavyrulewidth{.25ex}
\aboverulesep=0ex
\belowrulesep=.3ex
\centering
\caption{\textbf{Effect of the method used to extract sparse correspondences.} Feature matching is done with LightGlue~\cite{lindenberger2023lightglue} using different types of local features. We compare the performance of \method{} on the CVO (\textit{Final}) test set when fed with $N{=}1024$ correspondences extracted with different methods. For fair comparison, we do not do in-domain training (no specialization to any particular input), and we use the same densification model \camera{and the same sampling strategy (\ie, the one from Section~\ref{sec:overall}}) for all methods.} 
\small

\begin{tabular}{@{}L{1.cm}@{}L{2.45cm}@{}@{}C{2.9cm}@{}C{0.8cm}@{}C{1.2cm}@{}}
\toprule
\multicolumn{2}{@{}l@{}}{\normalsize Method} & EPE $\downarrow$ ({\scriptsize all / vis / occ}) & IoU $\uparrow$ & Time $\downarrow$ \\
\midrule
\multirow{3}{*}{\raisebox{-0.45\normalbaselineskip}[0pt][0pt]{\rotatebox[origin=c]{90}{\begin{tabular}{@{}c@{}} Feature \\ matching\end{tabular}}}} & SIFT~\cite{lowe2004distinctive} & 10.9 / 9.15 / 19.5 & 31.9 & 1.022 \\
& DISK~\cite{tyszkiewicz2020disk} & 10.9 / 8.81 / 21.2 & 32.9 & 0.362 \\
& ALIKED~\cite{zhao2023aliked} & 9.55 / 7.17 / 21.0 & 34.4 & \underline{0.282} \\
& SuperPoint~\cite{detone2018superpoint} & 8.76 / 6.20 / 21.0 & 36.7 & \textbf{0.244} \\
\midrule
\multirow{3}{*}{\raisebox{0.05\normalbaselineskip}[0pt][0pt]{\rotatebox[origin=c]{90}{\begin{tabular}{@{}c@{}} Point \\ tracking\end{tabular}}}} & PIPs++~\cite{zheng2023pointodyssey} & 7.30 / 4.92 / 19.5 & 48.0 & 4.102 \\
& TAPIR~\cite{doersch2023tapir} & \underline{4.00} / \underline{1.82} / \underline{14.3} & \underline{70.2} & 0.668 \\
& CoTracker~\cite{karaev2023cotracker} & \textbf{1.89} / \textbf{1.27} / \textbf{4.99} & \textbf{70.9} & 0.864 \\
\bottomrule 
\end{tabular}

\label{tab:cvo_corr_abl}
\end{table}

\begin{figure}
    \newcommand{\overview}[1]{%
    \begin{tikzpicture}
        \node[anchor=south west, inner sep=0, outer sep=0] (image) at (0,0) {\includegraphics[width=0.5\linewidth]{#1}};
        \draw [white, line width=2pt] (1.4,2.6) rectangle (2.1,3.3);
        \node[anchor=south west, text=black] at (2.1,2.6) {\setlength{\fboxsep}{2pt} \footnotesize \colorbox{white}{Target}};
        \node[anchor=south west, text=black] at (-0.05,0) {\setlength{\fboxsep}{2pt} \footnotesize \colorbox{white}{Source}};
        \node[anchor=south west, text=black] at (-0.05,-0.04) {};
    \end{tikzpicture}
    \hspace{-0.5em}}
    \newcommand{\collage}[2]{%
    \begin{tikzpicture}
        \node[anchor=south west, inner sep=0, outer sep=0] (image2) at (0,0) {\includegraphics[width=0.5\linewidth]{#1}};

        \begin{scope}
            \clip (0,0) -- (0.5\linewidth,0) -- (0.5\linewidth,0.5\linewidth) -- cycle; %
            \node[anchor=south west, inner sep=0, outer sep=0] (image1) at (0,0) {\includegraphics[width=0.5\linewidth]{#2}};
        \end{scope}

        \draw [white, line width=2.4pt] (0,0) -- (0.5\linewidth,0.5\linewidth);

        \node[anchor=north west, text=black] at (0,0.5\linewidth) {\setlength{\fboxsep}{2pt} \footnotesize \colorbox{white}{CoTracker}};
        \node[anchor=south east, text=black] at (0.5\linewidth,0) {\setlength{\fboxsep}{2pt} \footnotesize \colorbox{white}{\method{}}};
        
        \draw [black, line width=2pt] (0.25\linewidth-30,0.25\linewidth-10) rectangle (0.25\linewidth+30,0.25\linewidth+10);
        
    \end{tikzpicture}
    \hspace{-0.5em}}
    \newcommand{\zoomin}[2]{%
    \begin{tikzpicture}
        \node[anchor=south west, inner sep=0, outer sep=0] (image1) at (0,0) {\includegraphics[width=0.5\linewidth-2pt]{#1}};
        \draw [black, inner sep=2pt, line width=2pt] (image1.south west) rectangle (image1.north east);

        \node[anchor=south west, text=black] at (-0.05,0) {\setlength{\fboxsep}{2pt} \footnotesize \colorbox{white}{#2}};
        \node[anchor=south west, text=black] at (-0.1,0) {};
        
    \end{tikzpicture}
    \hspace{-0.5em}}

	\setlength\tabcolsep{1.5pt}
	\renewcommand{\arraystretch}{0.4}
	\small
    \centering
    \setlength{\fboxsep}{0pt}
    \setlength{\fboxrule}{1pt}
    \hspace{-0.75em}
    \begin{tabular}{@{}C{0.99\linewidth}@{}}
    \begin{tabular}{@{}cc@{}}
        \overview{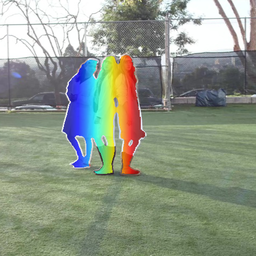} &
        \collage{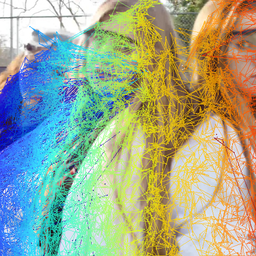}{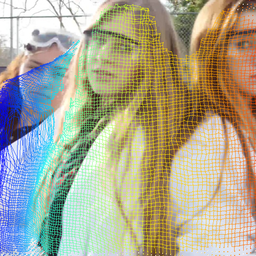} \\
        \zoomin{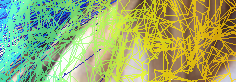}{CoTracker} & \hspace{-0.1em}\zoomin{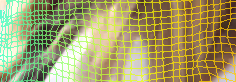}{\method{}}
        
	\end{tabular}
    \end{tabular}
	\caption{\textbf{Preservation of local structure.} We compare CoTracker and \method{} on the ``\textit{Lab-coat}'' video from the DAVIS dataset. We represent foreground points in the source frame ($s{=}1$) using distinctive colors, and visualize predicted tracks for the target frame ($t{=}40$) with the same colors: we show the mesh obtained by connecting each point at time $t$ to its cardinal neighbors from time $s$. \method{} preserves local structure much better than CoTracker.} %
	\label{fig:tap_struct}
\end{figure}

\begin{table}
\setlength\heavyrulewidth{.25ex}
\aboverulesep=0ex
\belowrulesep=.3ex
\centering
\caption{\textbf{Ablations of the core components of our approach.} We remove components one at a time and rank them from the least to the most significant in terms of flow and mask reconstruction quality on the CVO (\textit{Final}) test set. We use $N{=}1024$ input tracks.}
\small

\begin{tabular}{@{}L{4.7cm}@{}@{}C{2.8cm}@{}C{0.8cm}@{}}
\toprule
{\normalsize Method} & EPE $\downarrow$ ({\scriptsize all / vis / occ}) & IoU $\uparrow$ \\
\midrule
\textit{\Method{}} (\method{}) & \textbf{1.43} / \textbf{0.85} / \textbf{4.29} & \textbf{79.7} \\
- \camera{No use of visibility info from tracks} & \underline{1.46} / \underline{0.87} / \underline{4.31} & {78.9} \\
- No motion-based sampling of tracks & {1.50} / {0.90} / {4.45} & \underline{79.2} \\
- Patch size of $8$ instead of $4$ & 1.60 / 1.02 / 4.45 & 76.8 \\
- No in-domain training & 1.90 / 1.27 / 5.00 & 70.8 \\
- No track-based estimates & 2.88 / 1.79 / 7.89 & 57.2 \\
- No optical-flow refinement & 3.19 / 2.48 / 6.79 & 54.5 \\
\bottomrule 
\end{tabular}

\label{tab:cvo_abl}
\end{table}

\subsection{Ablation studies}

\paragraph{Effect of the method used to extract sparse correspondences.}
We compare the performance of \method{} when different methods are used to extract input correspondences in Table~\ref{tab:cvo_corr_abl}.
We explore local feature matching techniques which produce correspondences between pairs of images.
Although they allow fast computations, such methods may produce incorrect matches when parts of different objects present important similarities~\cite{cai2023doppelgangers}. 
Moreover, they do not handle occlusions, with points required to be visible in both images, and are not very robust to motion blur.
We found that resulting correspondences are not well distributed, \ie, essentially on the background, as illustrated in Appendix~\ref{sec:cvo_corr_qual}.
Conversely, point tracking methods solve this issue by leveraging temporal information, leading to notable improvements.
Among these methods, we have opted for the efficient variant of CoTracker~\cite{karaev2023cotracker} since it provides superior motion reconstructions at a reasonable processing speed. 

\paragraph{Ablations of the core components of our approach} are in Table~\ref{tab:cvo_abl}, with one component removed at a time.
\camera{Using visibility information from tracks slightly improves motion predicition (EPE) and visibility prediction (IoU).}
We compare randomly sampling tracks to our motion-based strategy and observe that the latter yields more informative tracks.
We find that changing the patch size for flow refinement from $P{=}8$ to $P{=}4$, effectively doubling the resolution of the features, improves performance. A more detailed analysis is in Appendix~\ref{sec:cvo_patch}.
In-domain training, \ie, specializing our densification model to noisy estimates from a specific point tracking model, as opposed to training with ground-truth tracks as input, is also helpful.
Moreover, relying solely on the optical flow component of our approach for a pair of source and target frames, without incorporating track-based estimates, results in much worse performance.
Similarly, maintaining the same number of input tracks but omitting optical flow refinement does not yield satisfactory results.
\section{Conclusion}
\label{sec:conclusion}

We have introduced \method{}, an approach for dense motion estimation which unifies optical flow and point tracking techniques, effectively leveraging the strengths from both: reaching the accuracy of the latter with the speed and spatial coherence of the former. 
Like any other approach, it, of course, may fail due to extreme occlusions, fast motions, or rapid changes in appearance.
\method{} is agnostic to the choice of a specific point tracking algorithm, so future advances in this field will directly benefit our approach. We believe that the efficiency of \method{} holds the potential to drive substantial progress across various downstream applications.
\section*{Acknowledgements}

This work was granted access to the HPC resources of IDRIS under the allocation 2021-AD011012227R2 made by GENCI. It was funded in part by the French government under management of Agence Nationale de la Recherche as part of the ``Investissements d’avenir'' program, reference ANR-19-P3IA-0001 (PRAIRIE 3IA Institute), and the ANR project VideoPredict, reference ANR-21-FAI1-0002-01.
JP was supported in part by the Louis Vuitton/ENS chair in artificial intelligence and a Global Distinguished Professorship at the Courant Institute of Mathematical Sciences and the Center for Data Science at New York University.
{
    \small
    \bibliographystyle{ieeenat_fullname}
    \bibliography{main}
}
\clearpage
\newpage
\onecolumn
\appendix
\begin{center}
{\Large\bfseries Dense Optical Tracking: Connecting the Dots}

\vspace{0.5em}

{\it \large Appendix}

\vspace{2em}
\end{center}

\section{Detailed architecture}
\renewcommand{\thefigure}{A}
\label{sec:arch}

Implementation details for our optical flow module are presented in Figure~\ref{fig:arch}. The two major differences with the original RAFT network architecture~\cite{teed2020raft} are \raisebox{0.1em}{\setlength{\fboxsep}{0.5pt}\setlength{\fboxrule}{1.2pt}\fcolorbox{blue}{white}{\footnotesize highlighted}} in the figure. First, we use a stride 1 instead of 2 in the first convolutional layer of the frame encoder ($\mathcal{E}_X$), such that the spatial resolution is decreased by a factor $P{=}4$ instead of $P{=}8$. We find that this has a significant impact on performance, see ablation studies in Table~\ref{tab:cvo_abl} of the paper and also additional results in Appendix~\ref{sec:cvo_patch}. Second, our approach not only predicts a dense flow field, but also a visibility mask. We thus adapt the encoder ($\mathcal{E}$) and add a new decoder ($\mathcal{D}_M$) to take into account this new modality in the refinement process. 

\begin{figure}[ht]
\centering
\setlength\tabcolsep{5.pt}
\begin{tabular}{@{}cc@{}}
    \adjustbox{valign=b}{\begin{subfigure}[t]{0.7\textwidth}
        \centering
        \includegraphics[width=\linewidth]{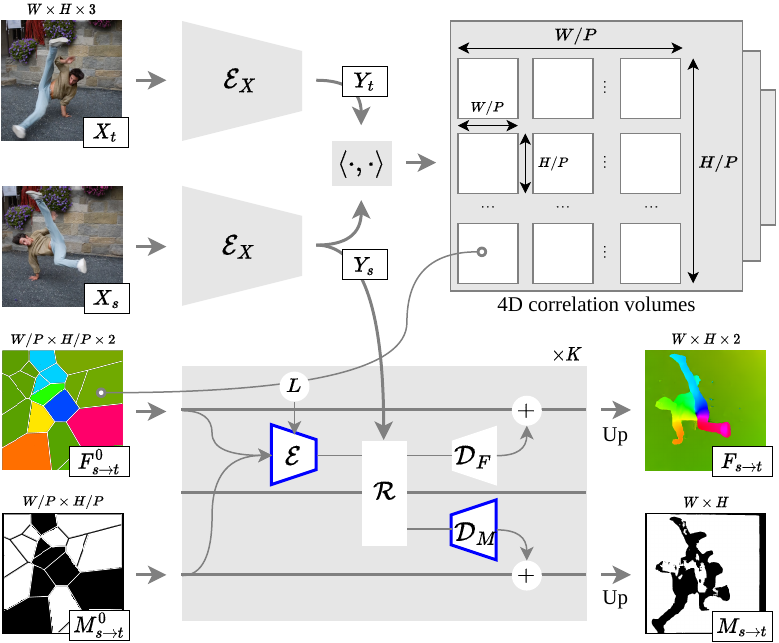}
        \caption{Overview.}
    \end{subfigure}} &
    \adjustbox{valign=b}{\begin{tabular}{@{}c@{}}
    \begin{subfigure}[t]{0.27\textwidth}
        \footnotesize
        \begin{tabular}{@{}C{1.4cm}@{}C{1.2cm}@{}C{1.2cm}@{}C{0.8cm}@{}}
            \multicolumn{4}{@{}c@{}}{\cellcolor{mygray}Residual encoder} \\
            Conv$7\times7$ & 3 & 64 & {\setlength{\fboxsep}{0.5pt}\setlength{\fboxrule}{1.2pt}%
    \fcolorbox{blue}{white}{~~~1~~~}} \\
            Res$3\times3$ & 64 & 64 & 1 \\
            Res$3\times3$ & 64 & 96 & 2 \\
            Res$3\times3$ & 96 & 128 & 2 \\
            Conv$1\times1$ & 128 & 256 & 1 \\
        \end{tabular}
        \caption{Frame encoder ($\mathcal{E}_X$).}
    \end{subfigure} \\ \\[0.5em]
    \begin{subfigure}[t]{0.27\textwidth}
        \footnotesize
        \begin{tabular}{@{}C{1.4cm}@{}C{1.2cm}@{}C{1.2cm}@{}C{0.8cm}@{}}
            \multicolumn{4}{@{}c@{}}{\cellcolor{mygray}Flow and mask encoder} \\
            Conv$7\times7$ & 2{+}1 & 128 & 1 \\
            Conv$3\times3$ & 128 & 64 & 1 \\
            \multicolumn{4}{@{}c@{}}{\cellcolor{mygray}Correlation encoder} \\
            Conv$1\times1$ & 324 & 256 & 1 \\
            Conv$3\times3$ & 256 & 192 & 1 \\
            \multicolumn{4}{@{}c@{}}{\cellcolor{mygray}Combination of outputs from both} \\
            Conv$3\times3$ & 64+192 & 125 & 1 \\
        \end{tabular}
        \caption{Joint flow and mask encoder ($\mathcal{E}$).}
    \end{subfigure} \\ \\[0.5em]
    \begin{subfigure}[t]{0.27\textwidth}
        \footnotesize
        \begin{tabular}{@{}C{1.4cm}@{}C{1.2cm}@{}C{1.2cm}@{}C{0.8cm}@{}}
            \multicolumn{4}{@{}c@{}}{\cellcolor{mygray}Convolutional gated recurrent unit} \\
            GRU$5\times5$ & 128+256 & 128 & 1 \\
        \end{tabular}
        \caption{Recurrent unit ($\mathcal{R}$).}
    \end{subfigure} \\ \\[0.5em]
    \begin{subfigure}[t]{0.27\textwidth}
        \footnotesize
        \begin{tabular}{@{}C{1.4cm}@{}C{1.2cm}@{}C{1.2cm}@{}C{0.8cm}@{}}
            \multicolumn{4}{@{}c@{}}{\cellcolor{mygray}Flow / mask decoders} \\
            Conv$3\times3$ & 128 & 256 & 1 \\
            Conv$3\times3$ & 256 & 2/1 & 1 \\
            \midrule
            Operation & In. dim. & Out. dim. & Stride \\
        \end{tabular}
        \caption{Flow and mask decoders ($\mathcal{D}_F$ / $\mathcal{D}_M$).}
    \end{subfigure}\end{tabular}}
\end{tabular}
\caption{\textbf{Optical flow module.} Our approach builds upon RAFT~\cite{teed2020raft} and refines a dense flow field $F_{{s}\rightarrow{t}}$ and a visibility mask $M_{{s}\rightarrow{t}}$ from coarse estimates $F^0_{{s}\rightarrow{t}}$ and $M^0_{{s}\rightarrow{t}}$ using frames $X_s$ and $X_t$. We first use a frame encoder ($\mathcal{E}_X$), implemented as a convolutional neural network with residual connections that extracts features $Y_s$ and $Y_t$ in $\RR^{W/P \times H/P \times D}$ ($P{=}4$ and $D{=}256$ in practice) from input frames. We then compute the correlation between features for all pairs of source and target positions at different feature resolution levels, yielding a 4D correlation volume for each resolution level. More precisely, the size of the volume at the finest level is $W/P \times H/P \times W/P \times H/P$, and coarser levels are obtained by applying average pooling on the feature maps. The refinement process is composed of $K$ iterations of the following operations ($K{=}4$ in practice): A look-up operation ($L$) in the cost volumes using the flow estimate from the previous iteration; The joint encoding ($\mathcal{E}$) of the flow, mask and correlation information; A recurrent unit ($\mathcal{R}$) which is an adaptation of a GRU cell~\cite{cho2014properties} using convolutions; Two decoders ($\mathcal{D}_F$ and $\mathcal{D}_M$) to update the flow and the mask, respectively. We note that the model parameters are shared across all iterations. We finally upsample estimates from the K\textsuperscript{th} iteration, using for each position at the fine resolution a linear combination of the values for the 9 nearest neighbors at the coarse resolution. Please refer to RAFT~\cite{teed2020raft} for further details.}

\label{fig:arch}
\end{figure}

\section{Interpolation strategy}
\renewcommand{\thefigure}{B}
\label{sec:inter}

Let $N$ represent the number of initial tracks, and $(H, W)$ denote the spatial resolution after interpolating these tracks. In a naive implementation of nearest-neighbor interpolation, the memory complexity grows quadratically in $NHW$, as it requires computing the distance between every pixel and every track. This can lead to significant memory issues, particularly when dealing with high resolutions.
To address this challenge, we propose an efficient implementation relying on PyTorch3D~\footnote{\url{https://github.com/facebookresearch/pytorch3d/}} which comes with custom CUDA kernels, specifically optimized for this kind of operations.

\section{Space-time visualizations of motion}
\renewcommand{\thefigure}{C}
\label{sec:tap_space_time}

\begin{figure}[ht]
	\setlength\tabcolsep{1.pt}
	\renewcommand{\arraystretch}{0.25}
	\small
    \centering
    \setlength{\fboxsep}{0pt}
    \setlength{\fboxrule}{1pt}
	\begin{tabular}{@{}ccc@{}}
    \includegraphics[width=0.56\linewidth]{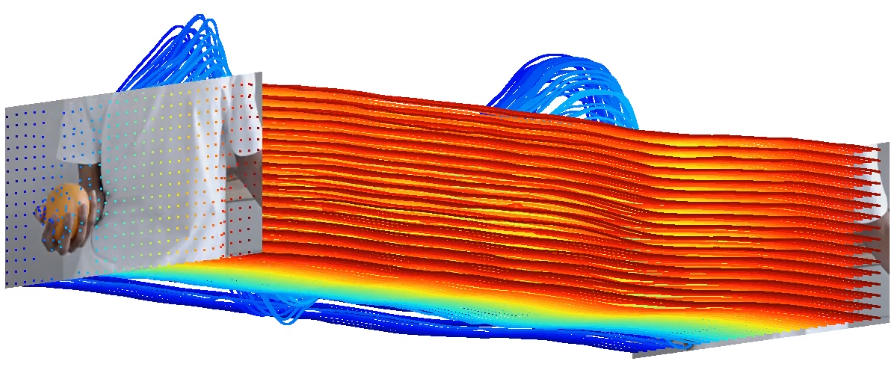} &
    \includegraphics[width=0.26\linewidth]{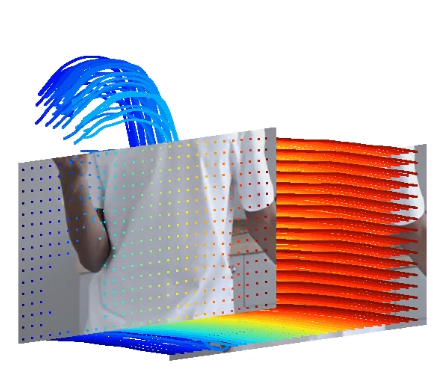} & \includegraphics[width=0.17\linewidth]{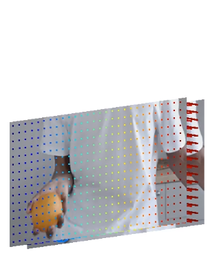} \\[0.5em]
    \multicolumn{3}{@{}c@{}}{(a) Throwing an orange} \\
    \includegraphics[width=0.56\linewidth]{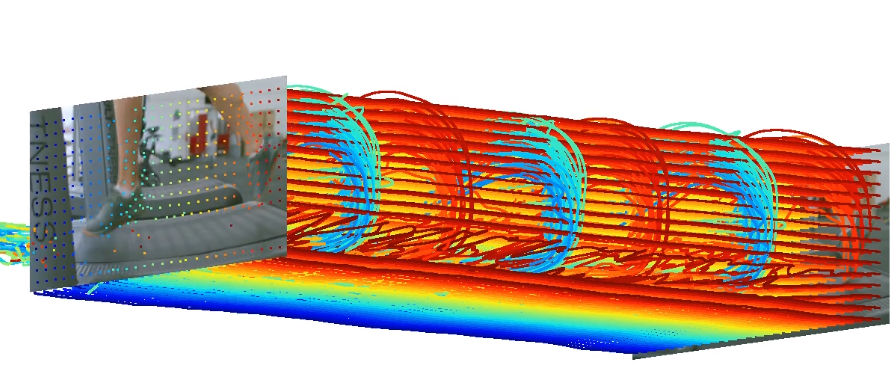} &
    \includegraphics[width=0.26\linewidth]{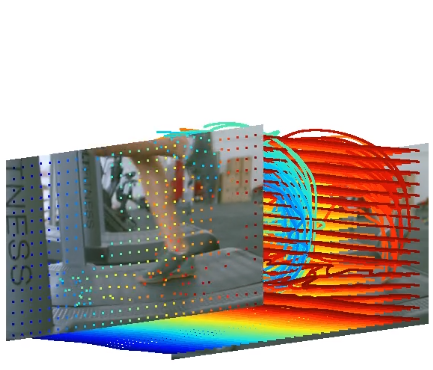} & \includegraphics[width=0.17\linewidth]{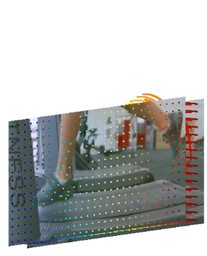} \\[0.5em]
    \multicolumn{3}{@{}c@{}}{(b) Running at the gym} \\
    \includegraphics[width=0.56\linewidth]{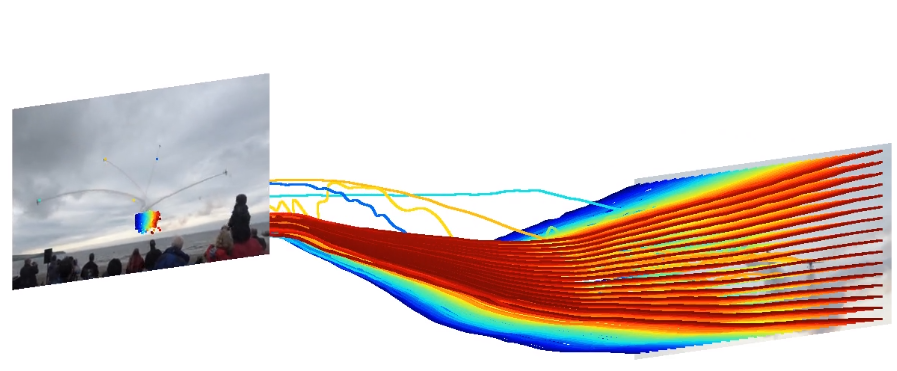} &
    \includegraphics[width=0.26\linewidth]{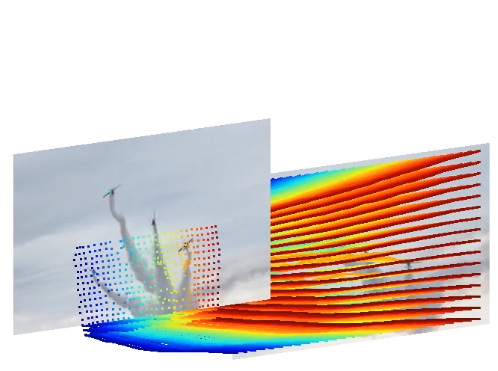} & \includegraphics[width=0.17\linewidth]{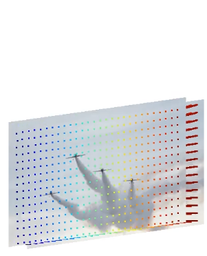} \\[0.5em]
    \multicolumn{3}{@{}c@{}}{(c) Airplane parade} \\
	\end{tabular}
	\vspace{-0.3em}
	\caption{\textbf{Space-time visualizations.} We track all pixels from the first frame of different videos, and show the trajectory for a subset of points, laid on a regular grid in the first frame. These trajectories are displayed in 3D, by using time as an additional dimension (progressing from right to left in the figure) alongside the two spatial dimensions. Each point on the grid is uniquely represented by a distinct color. Our method is able to track objects accurately, even when they go out of the frame multiple times like in video (a). \method{} is able to cope with intra-object occlusions, as in video (b) where the left leg occludes the right one repeatedly. It is also robust to extreme camera motions, as illustrated in video (c) where points, initially covering the whole frame, are concentrated in a small regions in the last frame. Please zoom in for details. See also the videos in the project webpage.} 
	\label{fig:tap_space_time}
\end{figure}

Qualitative results in Figure~\ref{fig:tap_space_time} show that \method{} produces accurate dense tracks even in challenging settings, such as when an object exits and re-enters the frame, when multiple objects occlude each other, or when facing extreme camera motions.

\clearpage

\section{Effect of the patch size on model performance and speed}
\label{sec:cvo_patch}
\renewcommand{\thefigure}{D}

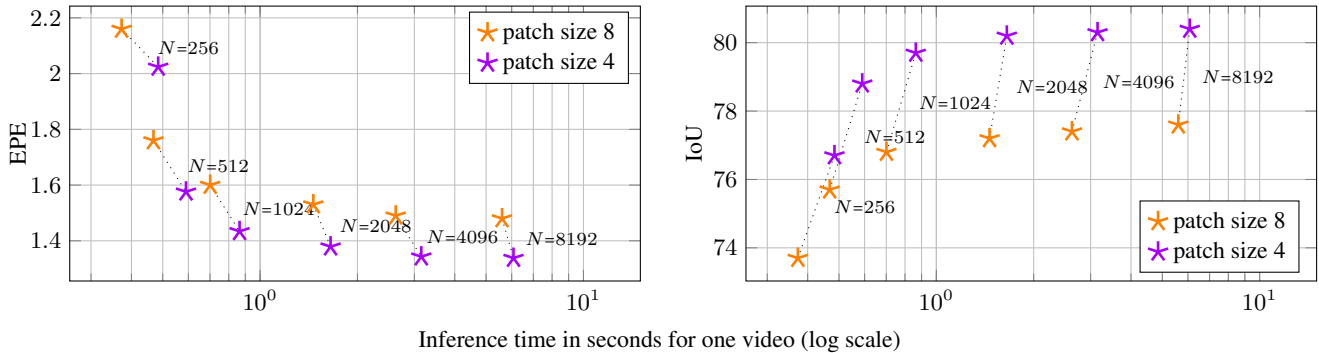
\begin{figure}[ht]
\centering
\small
\setlength\tabcolsep{7pt}
\begin{tabular}{@{}cc@{}}
\begin{tikzpicture}
    \begin{axis}[
        width=0.525\columnwidth,
        height=0.3\columnwidth,
        ylabel={EPE},
        xmode=log,
        xmax=15,
        grid=both,
        y label style={at={(axis description cs:0.1,0.5)}, anchor=south},
    ]
    
    \addplot[only marks, mark=star, mark options={draw=orange, fill=orange, scale=2, line width=1pt}, nodes near coords, point meta=explicit symbolic] table {
        Speed   Accuracy    Method
        0.373   2.16        \method{}
        0.468   1.76        \method{}
        0.701   1.60        \method{}
        1.461   1.53        \method{}
        2.627   1.49        \method{}
        5.602   1.48        \method{}
    };

    \addplot[only marks, mark=star, mark options={draw=mypurple, fill=mypurple, scale=2, line width=1pt}, nodes near coords, point meta=explicit symbolic] table {
        Speed   Accuracy    Method
        0.484   2.024       \method{}
        0.590   1.576       \method{}
        0.864   1.434       \method{}
        1.652   1.379       \method{}
        3.152   1.343       \method{}
        6.078   1.338       \method{}
    };

    \draw[dotted] (axis cs: 0.373, 2.16) -- (axis cs: 0.484, 2.024) node[midway, right] {\scriptsize ~~$N{=}256$};
    \draw[dotted] (axis cs: 0.468, 1.76) -- (axis cs: 0.590, 1.576) node[midway, right] {\scriptsize ~~$N{=}512$};
    \draw[dotted] (axis cs: 0.701, 1.60) -- (axis cs: 0.864, 1.434) node[midway, right] {\scriptsize ~~$N{=}1024$};
    \draw[dotted] (axis cs: 1.461, 1.53) -- (axis cs: 1.652, 1.379) node[midway, right] {\scriptsize ~~$N{=}2048$};
    \draw[dotted] (axis cs: 2.627, 1.49) -- (axis cs: 3.152, 1.343) node[midway, right] {\scriptsize ~~$N{=}4096$};
    \draw[dotted] (axis cs: 5.602, 1.48) -- (axis cs: 6.078, 1.338) node[midway, right] {\scriptsize ~~$N{=}8192$};

    \legend{patch size 8, patch size 4}

    \end{axis}
    
\end{tikzpicture} &

\begin{tikzpicture}
    \begin{axis}[
        width=0.525\columnwidth,
        height=0.3\columnwidth,
        ylabel={IoU},
        xmode=log,
        xmax=15,
        grid=both,
        y label style={at={(axis description cs:0.1,0.5)}, anchor=south},
        legend pos=south east
    ]
    
    \addplot[only marks, mark=star, mark options={draw=orange, fill=orange, scale=2, line width=1pt}, nodes near coords, point meta=explicit symbolic] table {
        Speed   Accuracy    Method
        0.373   73.7        \method{}
        0.468   75.7        \method{}
        0.701   76.8        \method{}
        1.461   77.2        \method{}
        2.627   77.4        \method{}
        5.602   77.6        \method{}
    };

    \addplot[only marks, mark=star, mark options={draw=mypurple, fill=mypurple, scale=2, line width=1pt}, nodes near coords, point meta=explicit symbolic] table {
        Speed   Accuracy    Method
        0.484   76.7        \method{}
        0.590   78.8        \method{}
        0.864   79.7        \method{}
        1.652   80.2        \method{}
        3.152   80.3        \method{}
        6.078   80.4        \method{}
    };

    \draw[dotted] (axis cs: 0.373, 73.7) -- (axis cs: 0.484, 76.7) node[midway, right] {\scriptsize ~~$N{=}256$};
    \draw[dotted] (axis cs: 0.468, 75.7) -- (axis cs: 0.590, 78.8) node[midway, right] {\scriptsize ~~$N{=}512$};
    \draw[dotted] (axis cs: 0.701, 76.8) -- (axis cs: 0.864, 79.7) node[midway, right] {\scriptsize ~~$N{=}1024$};
    \draw[dotted] (axis cs: 1.461, 77.2) -- (axis cs: 1.652, 80.2) node[midway, right] {\scriptsize ~~$N{=}2048$};
    \draw[dotted] (axis cs: 2.627, 77.4) -- (axis cs: 3.152, 80.3) node[midway, right] {\scriptsize ~~$N{=}4096$};
    \draw[dotted] (axis cs: 5.602, 77.6) -- (axis cs: 6.078, 80.4) node[midway, right] {\scriptsize ~~$N{=}8192$};

    \legend{patch size 8, patch size 4}

    \end{axis}
    
\end{tikzpicture} \\
\multicolumn{2}{@{}c@{}}{Inference time in seconds for one video (log scale)}
\end{tabular}

\caption{\textbf{Effect of the patch size} on flow reconstruction, occlusion prediction and speed on the CVO (\textit{Final}) dataset. We report the end point error (EPE) of flows, the intersection over union (IoU) of occluded regions, and the inference time (in seconds) when setting the number $N$ of initial tracks to different values in $[256,512,1024,2048,4096,8192]$.}
\label{fig:cvo_patch}
\end{figure}

We evaluate the impact of the patch size in the optical flow refinement module of our approach on the trade-off between performance and speed in Figure~\ref{fig:cvo_patch}. This involves setting the number $N$ of initial point tracks to different values. We find that having a patch size of $P{=}4$ instead of $P{=}8$ is almost always beneficial, with a great boost in performance at the cost of a small reduction in inference speed. 
We note that a similar study conducted in CoTracker~\cite{karaev2023cotracker} yields the same conclusions.
It is only for very small numbers of initial tracks that the larger patch size reaches better trade-offs, for example, with better performance in terms of EPE at a similar speed for $(P,N)=(8,512)$ compared to $(P,N)=(4,256)$.

\section{Effect of the number of tracks on initial and final motion estimates}
\renewcommand{\thefigure}{E}
\label{sec:tap_tracks}

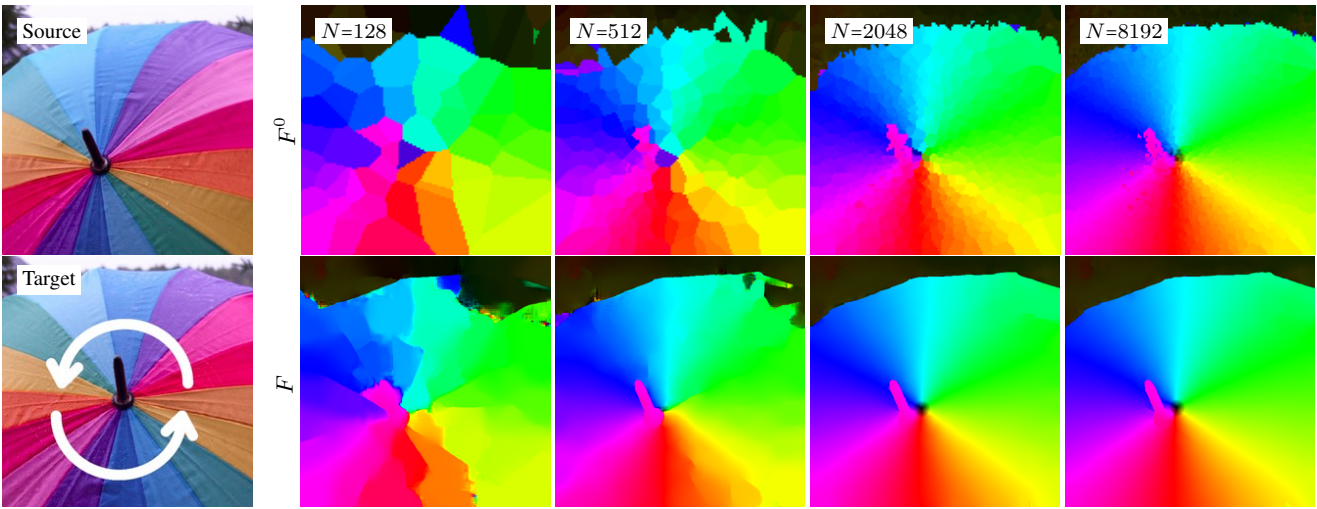
\begin{figure}[ht]
    \newcommand{\withName}[2]{%
    \begin{tikzpicture}
        \node[anchor=south west, inner sep=0, outer sep=0] (image) at (0,0) {#1};
        \node[anchor=south west, text=black] at (0,2.7) {\setlength{\fboxsep}{2pt} \footnotesize \colorbox{white}{#2}};
    \end{tikzpicture}
    \hspace{-0.3em}}
	\setlength\tabcolsep{1.pt}
	\renewcommand{\arraystretch}{0.25}
	\small
    \centering
    \setlength{\fboxsep}{0pt}
    \setlength{\fboxrule}{1pt}
	\begin{tabular}{@{}cccccc@{}}
	\withName{\includegraphics[width=0.19\linewidth]{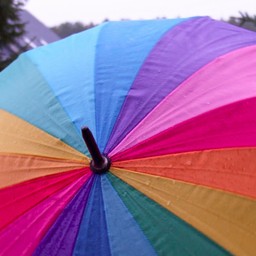}}{Source} & 
    ~~~\raisebox{3.9\normalbaselineskip}[0pt][0pt]{\rotatebox[origin=c]{90}{$F^0$}} & 
    \withName{\includegraphics[width=0.19\linewidth]{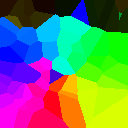}}{$N{=}128$} & 
    \withName{\includegraphics[width=0.19\linewidth]{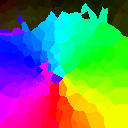}}{$N{=}512$} &
    \withName{\includegraphics[width=0.19\linewidth]{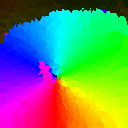}}{$N{=}2048$} & 
    \withName{\includegraphics[width=0.19\linewidth]{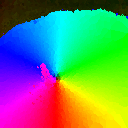}}{$N{=}8192$} \\
    \withName{\includegraphics[width=0.19\linewidth]{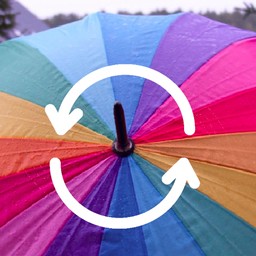}}{Target} & 
    ~~~\raisebox{3.9\normalbaselineskip}[0pt][0pt]{\rotatebox[origin=c]{90}{$F$}} &
    \includegraphics[width=0.19\linewidth]{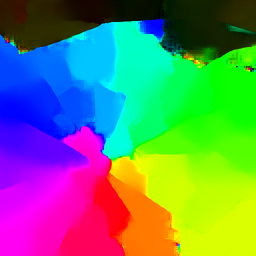} & 
    \includegraphics[width=0.19\linewidth]{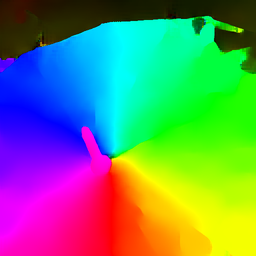} &
    \includegraphics[width=0.19\linewidth]{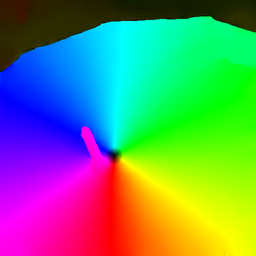} & 
    \includegraphics[width=0.19\linewidth]{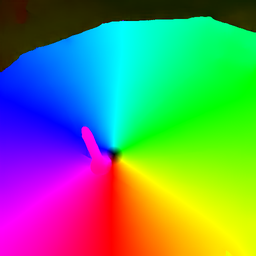} \\
	\end{tabular}
	\caption{\textbf{Effect of the number of tracks.} We visualize the initial flow between source and target frames ($F^0$) obtained by applying nearest-neighbor interpolation to $N$ tracks extracted with an off-the-shelf model~\cite{karaev2023cotracker}, and the final flow ($F$) refined by our approach. We represent motion directions using distinctive colors. We set the number $N$ of initial tracks to different values in $[128,512,2048,8192]$. The scene is composed of an umbrella undergoing a $\sim$180$^{\circ}$ rotation (see the white arrows superimposed on the target frame) and a small translation, which is the reason for the nonzero flow associated with the tip of the umbrella. }
	\label{fig:tap_tracks}
\end{figure}

We show in Figure~\ref{fig:tap_tracks} some qualitative samples for different numbers of initial tracks.
We see that final motion estimates produced by our method consistently improve over initial ones obtained through nearest-neighbor interpolation of tracks.
Our refinement process not only enhances spatial smoothness but also produces motions that better match object edges like the border between the umbrella and the background or the part that sticks out in the middle.

\section{Effect of the method used to extract sparse correspondences}
\renewcommand{\thefigure}{F}
\label{sec:cvo_corr_qual}

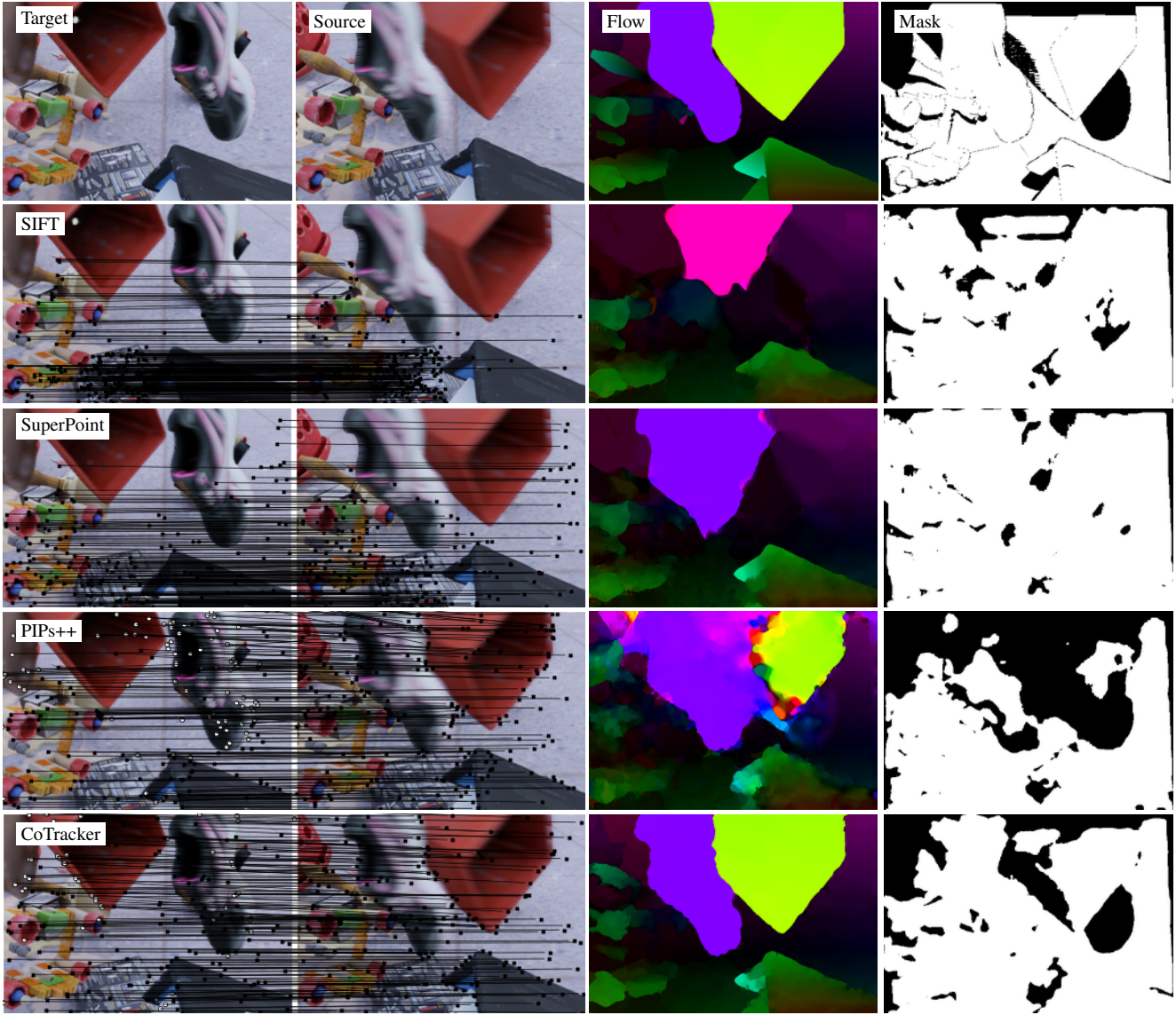
\begin{figure}[ht]
    \newcommand{\withName}[2]{%
    \begin{tikzpicture}
        \node[anchor=south west, inner sep=0, outer sep=0] (image) at (0,0) {#1};
        \node[anchor=south west, text=black] at (0,2.4) {\setlength{\fboxsep}{2pt} \footnotesize \colorbox{white}{#2}};
    \end{tikzpicture}
    \hspace{-0.3em}}
	\setlength\tabcolsep{1.pt}
	\renewcommand{\arraystretch}{0.25}
	\small
    \centering
    \setlength{\fboxsep}{0pt}
    \setlength{\fboxrule}{1pt}
	\begin{tabular}{@{}cccc@{}}
    
    \withName{\includegraphics[width=0.247\linewidth]{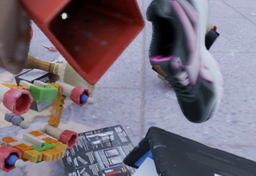}}{Target} & \withName{\includegraphics[width=0.247\linewidth]{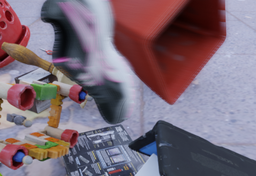}}{Source} &  \withName{\includegraphics[width=0.247\linewidth]{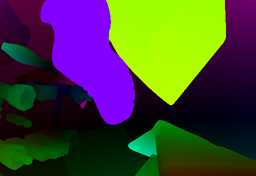}}{Flow} & \withName{\includegraphics[width=0.247\linewidth]{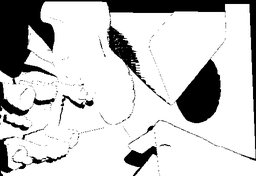}}{Mask} \\[1pt]
    
    \multicolumn{2}{@{}c@{}}{\withName{\includegraphics[width=0.498\linewidth]{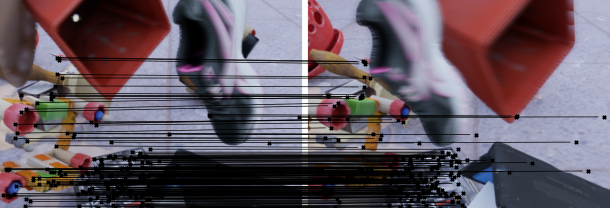}}{SIFT}}\hspace{-0.4em} & \includegraphics[width=0.247\linewidth]{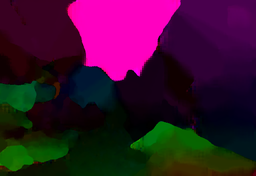} & \includegraphics[width=0.247\linewidth]{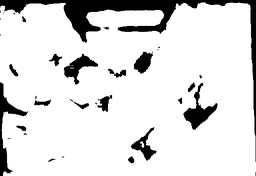} \\[0.5pt]
    
    \multicolumn{2}{@{}c@{}}{\withName{\includegraphics[width=0.498\linewidth]{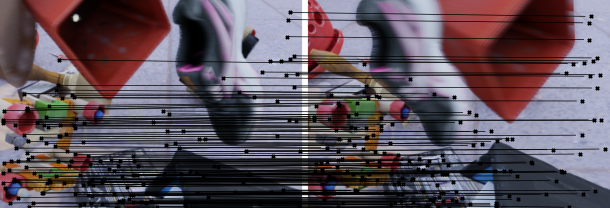}}{SuperPoint}}\hspace{-0.4em} & \includegraphics[width=0.247\linewidth]{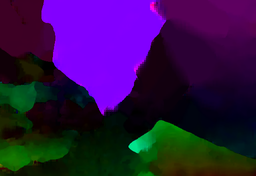} & \includegraphics[width=0.247\linewidth]{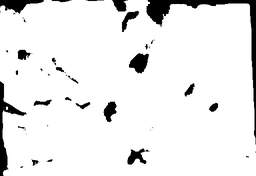} \\[1pt]

    \multicolumn{2}{@{}c@{}}{\withName{\includegraphics[width=0.498\linewidth]{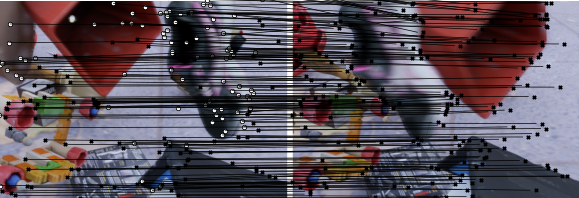}}{PIPs++}}\hspace{-0.4em} & \includegraphics[width=0.247\linewidth]{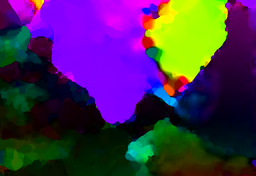} & \includegraphics[width=0.247\linewidth]{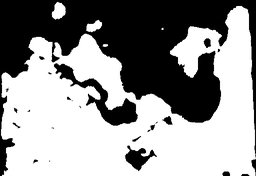} \\[1pt]

    \multicolumn{2}{@{}c@{}}{\withName{\includegraphics[width=0.498\linewidth]{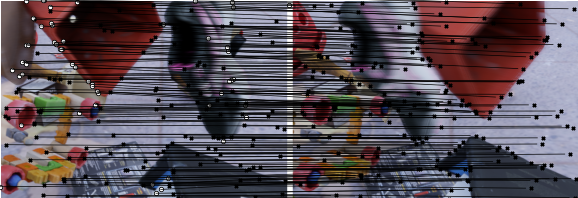}}{CoTracker}}\hspace{-0.4em} & \includegraphics[width=0.247\linewidth]{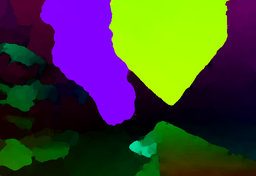} & \includegraphics[width=0.247\linewidth]{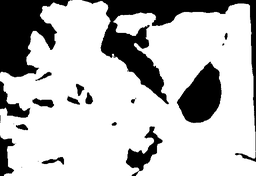} \\
    
	\end{tabular}
	\vspace{-0.3em}
	\caption{\textbf{Effect of the method used to extract sparse correspondences.} We show the flow and visibility mask produced by \method{} on the CVO (\textit{Final}) test set when fed with $N{=}1024$ correspondences from different methods. For clarity, we only show correspondences for $256$ pairs of points for each method (\setBold[0.5]$\times$\unsetBold: visible, \raisebox{-0.05\normalbaselineskip}[0pt][0pt]{\large\setBold[0.5]$\circ$\unsetBold}: occluded). For fair comparison, we use the same densification model for all methods and do not do in-domain training. We represent motion directions in the flow using distinctive colors and visible regions in the mask in white.} %
	\label{fig:cvo_corr_qual}
\end{figure}

We show in Figure~\ref{fig:cvo_corr_qual} correspondences between source and target frames obtained by various methods. Our approach refines dense motions (optical flow and visibility mask) from these correspondences.
Local feature matching methods, like SIFT~\cite{lowe2004distinctive} or SuperPoint~\cite{detone2018superpoint} with LightGlue~\cite{lindenberger2023lightglue}, excel in detecting salient feature points but struggle with textureless objects (\eg, the red pot) or regions with motion blur (\eg, the shoe).
The resulting flow for these objects is often far from the ground truth.
Moreover, as these methods only associate visible points, they offer limited assistance in predicting occlusions.
Point tracking methods such as PIPs++\cite{zheng2023pointodyssey} and CoTracker\cite{karaev2023cotracker} exhibit robust performance, predicting correspondences across the entire image. They consider all frames between source and target time steps, tracking points even under occlusion, providing informative estimates for our method. Among these, we adopt CoTracker as it yields higher quality correspondences. %

\clearpage

\section{Robustness to appearance changes}
\renewcommand{\thefigure}{G}
\label{sec:tap_changes}

\camera{We make DOT robust to appearance changes by training on data featuring shadows and motion blur effects. CoTracker and RAFT, trained on the same data, are less robust to such changes, as shown in Figure~\ref{fig:changes}.}

\newcommand{\withName}[2]{%
    \hspace{-0.5em}
    \begin{tikzpicture}
        \node[anchor=south west, inner sep=0, outer sep=0] (image) at (0,0) {#1};
        \node[anchor=south west, text=black] at (-0.1,1.35) {\setlength{\fboxsep}{2pt} \scriptsize \colorbox{white}{#2}};
    \end{tikzpicture}
    \hspace{-0.3em}}
    
\begin{figure}[ht]
    \footnotesize
    \centering
    \hspace{0.2em}\withName{\includegraphics[width=0.24\linewidth]{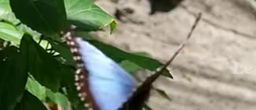}}{Source}
    \includegraphics[width=0.24\linewidth]{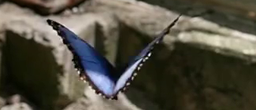}
    \includegraphics[width=0.24\linewidth]{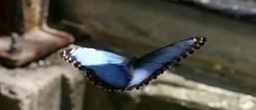}
    \includegraphics[width=0.24\linewidth]{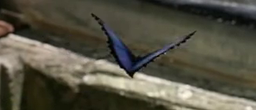}
    \withName{\includegraphics[width=0.24\linewidth]{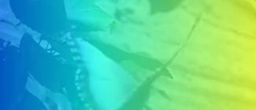}}{RAFT}
    \includegraphics[width=0.24\linewidth]{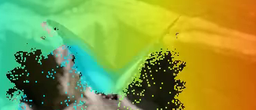}
    \includegraphics[width=0.24\linewidth]{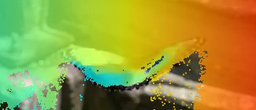}
    \includegraphics[width=0.24\linewidth]{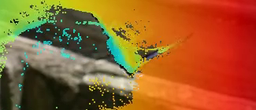}
    \withName{\includegraphics[width=0.24\linewidth]{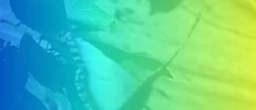}}{CoTracker}
    \includegraphics[width=0.24\linewidth]{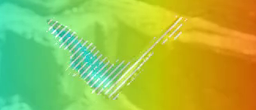}
    \includegraphics[width=0.24\linewidth]{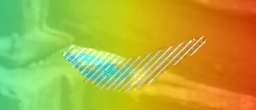}
    \includegraphics[width=0.24\linewidth]{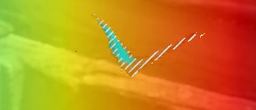}
    \withName{\includegraphics[width=0.24\linewidth]{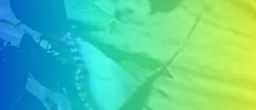}}{DOT}
    \includegraphics[width=0.24\linewidth]{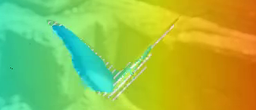}
    \includegraphics[width=0.24\linewidth]{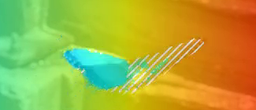}
    \includegraphics[width=0.24\linewidth]{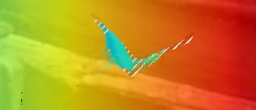}\hspace{-0.2em}
    \vspace{-1em}
    \caption{\textbf{Robustness to appearance changes.} \camera{We track all points in the first frame of the ``butterfly'' video
from the DAVIS dataset. RAFT loses part of the wing. CoTracker struggles with visibility.}}
    \label{fig:changes}
\end{figure}

\section{Data curation pipeline}
\renewcommand{\thefigure}{H}
\label{sec:cvo_curation}

\begin{figure}[ht]
	\setlength\tabcolsep{1.pt}
	\renewcommand{\arraystretch}{0.5}
	\small
    \centering
	\begin{tabular}{@{}C{0.333\linewidth}@{}C{0.333\linewidth}@{}C{0.333\linewidth}@{}}
    \multicolumn{3}{@{}c@{}}{\includegraphics[width=\linewidth]{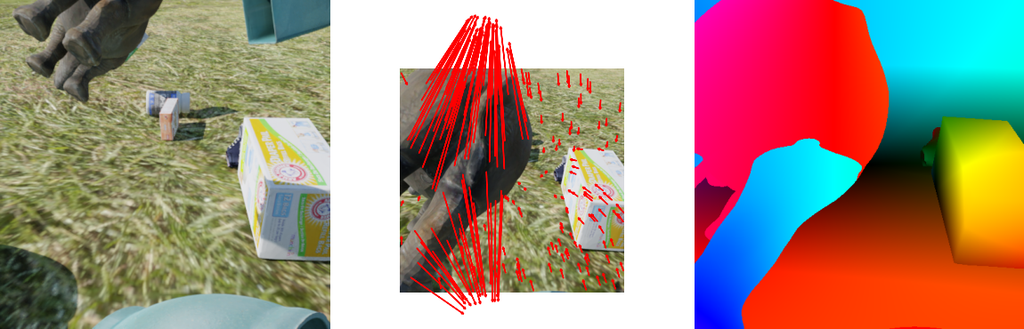}}  \\     
    Target frame & Source frame with motion vectors & Optical flow \\
	\end{tabular}
	\caption{\textbf{Data curation pipeline on the CVO dataset.} We identify samples with corrupted ground truth by visualizing source and target frames and the corresponding optical flow map side-by-side. In most cases, objects whose flow is incorrect have motion vectors pointing to very different directions. So we ease the verification process by showing a few motion vectors on top of the source frame. The corrupted part in the example presented here is the right hind leg of the toy.}
	\label{fig:cvo_curation}
\end{figure}

We have found that some videos from the CVO test set have erroneous optical flow ground truths due to objects being too close to the simulated camera. We have thus performed a systematic visual test for all the videos by showing simultaneously source and target frames, and the corresponding optical flow maps, as illustrated in Figure~\ref{fig:cvo_curation}. We have identified 25 corrupted samples out of more than 500 videos in the \textit{Clean} and \textit{Final} test sets but have not found any in our \textit{Extended} set. We filter out these few problematic samples when comparing different methods in our experiments. 

\end{document}